\documentclass[10pt,twocolumn,letterpaper]{article}

\usepackage{cvpr}              % To produce the CAMERA-READY version
\usepackage{graphicx}
\usepackage{amsmath}
\usepackage{amssymb} % for \mathbb
\usepackage{algorithm}
\usepackage{makecell}
\usepackage[noend]{algpseudocode}
\usepackage{booktabs}
\usepackage{multirow}

\definecolor{cvprblue}{rgb}{0.21,0.49,0.74}
\usepackage[pagebackref,breaklinks,colorlinks,allcolors=cvprblue]{hyperref}

\usepackage{colortbl}
\usepackage{xcolor}
\usepackage{multirow}

\title{
Attention-space Contrastive Guidance\\for Efficient Hallucination Mitigation in LVLMs}
\def\paperID{43464} % *** Enter the Paper ID here
\def\confName{CVPR}
\def\confYear{2026}

\author{
Yujin Jo,
Sangyoon Bae,
Taesup Kim$^\dagger$\\[2mm]
Seoul National University \\}

\begin{document}
\maketitle

\begingroup
\renewcommand{\thefootnote}{\fnsymbol{footnote}}
\setcounter{footnote}{2} % 1 -> dagger in fnsymbol sequence
\footnotetext{Corresponding author.}
\endgroup

\begin{abstract}
Hallucinations in large vision--language models (LVLMs) often arise when language priors dominate over visual evidence, leading to object misidentification and visually inconsistent descriptions.
We address this problem by framing hallucination mitigation as contrastive guidance that steers generation toward visually grounded and semantically faithful text.
We propose Attention-space Contrastive Guidance (ACG), a training-free, single-pass method that operates directly in self-attention layers, where hallucination-inducing cross-modal biases emerge.
ACG constructs both image-conditioned and approximate text-only attention paths within a single forward pass, enabling efficient guidance before errors accumulate at the output layer.
Because this masking-based surrogate can introduce approximation bias, we further apply a lightweight orthogonal projection that suppresses components aligned with the text-only path, yielding a more visually grounded correction.
Experiments on CHAIR and POPE show that ACG improves faithfulness over existing training-free baselines while maintaining caption quality, reducing latency by up to $2\times$ compared to multi-pass contrastive decoding methods.
\end{abstract}

% \begin{abstract}
% Hallucinations in large vision–language models (LVLMs) often arise when language priors dominate over visual evidence, causing object misidentification and visually inconsistent descriptions.
% We address this issue by framing hallucination mitigation as contrastive guidance, steering generation toward visually grounded and semantically faithful text. 
% This approach regulates the model’s internal behavior by reducing over-dependence on language priors and contrasting visually grounded with language-only representations.
% We propose Attention-space Contrastive Guidance (ACG), a single-pass mechanism that operates within self-attention layers to construct both vision–language and language-only attention paths in a single forward computation.
% This integration enables computationally efficient guidance directly embedded in the model’s representation contextualization. 
% To correct approximation bias introduced by the single-pass formulation, we further apply an orthogonalized correction that removes components aligned with the language-only path, selectively amplifying visual contributions.
% Experiments on the CHAIR and POPE benchmarks show that ACG achieves state-of-the-art faithfulness and caption quality while significantly reducing computational cost. Our method establishes a principled and efficient alternative, reducing latency by up to 2× compared to prior contrastive decoding methods that require multiple forward passes.

% \end{abstract}

\section{Introduction}
\label{sec:intro}

% \begin{figure}[t!]
%     \centering
%     \includegraphics[width=0.95\columnwidth]{figure/main_fig.png} 
%     \caption{\textbf{Framework.} Attention Contrastive Guidance.}
%     \label{fig:main}
% \end{figure}
\begin{figure*}[t!]
    \centering
    \includegraphics[width=0.95\textwidth]{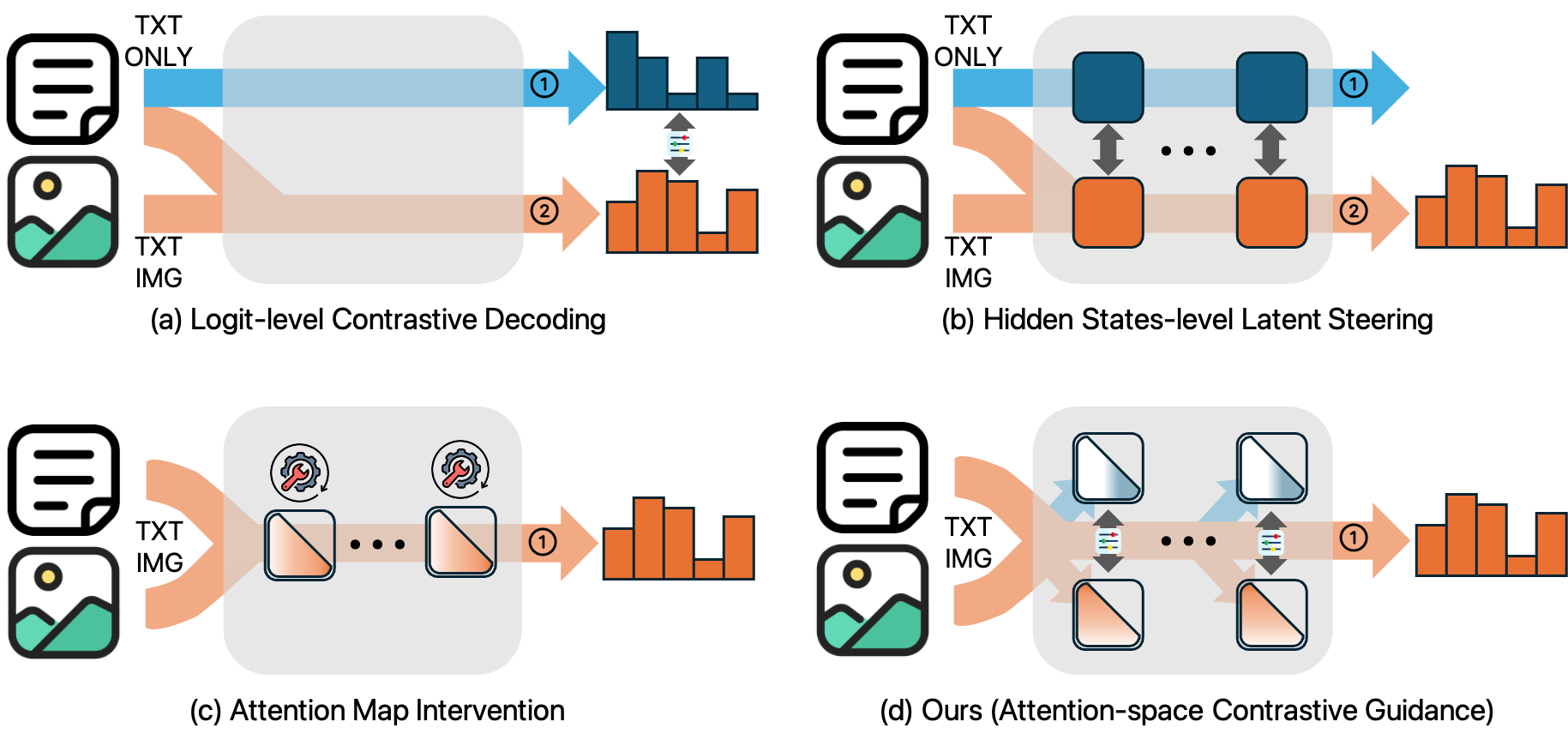} 
    \caption{\textbf{Comparison of inference-time strategies for mitigating LVLM hallucinations.} (a) Logit-level contrastive decoding, (b) hidden-state-level latent steering, (c) attention map intervention, and (d) the proposed Attention-space Contrastive Guidance (ACG).}
\label{fig:main}
\end{figure*}

Large vision--language models (LVLMs) have recently shown impressive performance on a wide range of multimodal tasks~\cite{llava1_5, dai2023instructblipgeneralpurposevisionlanguagemodels, li2023blip2bootstrappinglanguageimagepretraining, alayrac2022flamingovisuallanguagemodel, zhu2023minigpt4enhancingvisionlanguageunderstanding, bai2023qwenvlversatilevisionlanguagemodel}. These tasks include open-ended visual question answering, captioning, instruction following, and tool use. By combining strong visual encoders with powerful large language models, LVLMs can describe complex scenes, follow multimodal instructions, and reason over images using natural language.

Despite this rapid progress, however, LVLMs still suffer from a critical failure mode: hallucination. In such cases, the model generates text that is inconsistent with the visual evidence, for example by confidently describing objects that are not present in the image~\cite{chair, pope, mmhal, kaul2025throneobjectbasedhallucinationbenchmark}. This problem substantially undermines the reliability and trustworthiness of LVLMs. It is especially concerning in safety-critical applications such as medical imaging, autonomous driving, and robotics, where image-grounded reasoning is essential~\cite{zuo2025medhallbenchnewbenchmarkassessing}.

At a high level, hallucinations often arise when the model over-relies on language priors acquired during large-scale text pre-training and under-utilizes the actual visual evidence~\cite{bai2025hallucinationmultimodallargelanguage}. Instead of strictly conditioning on the input image, the model may fill in plausible but unobserved objects based on co-occurrence statistics. From this perspective, hallucinations can be viewed as a failure of controlled generation: the output is not sufficiently constrained by the visual condition and drifts toward language-only biases. One way to mitigate this behavior is to modify the model itself, for example through architectural changes or additional fine-tuning on hallucination-focused data. Recent works follow this direction by aligning LVLMs with human or synthetic preferences through RLHF or contrastive learning~\cite{mmhal, jiang2024hallucinationaugmentedcontrastivelearning, wu2025mitigatinghallucinationslargevisionlanguage}. While effective, such approaches are costly and inflexible, as they require parameter access and expensive retraining on carefully constructed supervision.

This cost and inflexibility have motivated training-free, inference-time methods that steer a fixed LVLM without updating its parameters. Closest to our work are logit-level guidance and contrastive decoding methods~\cite{pai, vcd, park2025secondmitigatingperceptualhallucination, icd}. These methods compare logits under image-conditioned and text-only (or weakened-image) inputs to penalize language-biased continuations and promote outputs that are better aligned with the visual condition. However, they act only on the final logits, after the model has already formed intermediate cross-modal representations, and therefore provide only a post-hoc correction. They also often require multiple forward passes to obtain a reference signal, which increases decoding latency. Recent methods have also explored attention-level interventions~\cite{ jiang2025devilsmiddlelayerslarge, adhh, chen2025spatialreasoninghardvlms, zhu2025mitigatingobjecthallucinationslarge, pai, jung2025visualattentionfadesselective, kang2025toldvisualattentionsink}. While promising, many of these approaches still rely on extra passes or additional steering rules, rather than providing a unified, single-pass mechanism for image-grounded control inside the model.

In this paper, we propose \emph{Attention-space Contrastive Guidance} (ACG), a training-free, single-pass guidance mechanism that operates directly within the self-attention layers of an LVLM. Instead of applying a single global correction at the output layer, ACG performs contrastive guidance in attention space by constructing image-conditioned and approximate text-only attention paths within a single forward pass, and using their difference to steer generation toward visual evidence as decoding unfolds. We further characterize the approximation bias introduced by the masking-based construction of the text-only path, and mitigate it with an orthogonalized correction that suppresses components aligned with the text-only path. This attention-space formulation enables fine-grained control over cross-modal alignment while remaining computationally efficient, allowing ACG to achieve comparable or better hallucination reduction than prior multi-pass contrastive decoding methods at substantially lower cost.

Our main contributions are as follows:
\begin{itemize}
    \item We formulate LVLM hallucination mitigation as contrastive guidance in attention space, and instantiate it as Attention-space Contrastive Guidance (ACG), a training-free, single-pass method that constructs image-conditioned and approximate text-only attention paths within each attention layer to correct cross-modal biases during decoding.
    \item To compensate for the approximation bias induced by the masking-based construction of the text-only path, we introduce an orthogonalized correction that suppresses components aligned with the text-only path, yielding a more visually grounded guidance signal.
    \item We demonstrate that ACG consistently improves faithfulness on standard hallucination benchmarks, including CHAIR and POPE, while achieving comparable or better hallucination reduction with up to 2$\times$ lower inference latency than prior training-free logit-level guidance baselines.
\end{itemize}

\section{Related Work}
\label{sec:related_work}

\subsection{LVLMs and Hallucinations}
Large vision--language models (LVLMs) have shown remarkable capabilities in integrating vision and language and have been evaluated on a wide range of multimodal benchmarks~\cite{llava1_5, dai2023instructblipgeneralpurposevisionlanguagemodels, li2023blip2bootstrappinglanguageimagepretraining, alayrac2022flamingovisuallanguagemodel, zhu2023minigpt4enhancingvisionlanguageunderstanding, bai2023qwenvlversatilevisionlanguagemodel}. 
However, their practical deployment is hindered by hallucinations, where models generate text that is inconsistent with the visual input. This phenomenon, ranging from describing nonexistent objects to misstating attributes or relations, is often attributed to an imbalance in which the model's powerful language priors override the actual visual evidence~\cite{chair, pope, kaul2025throneobjectbasedhallucinationbenchmark, petryk2024alohanewmeasurehallucination, bai2025hallucinationmultimodallargelanguage}.

Mitigating this failure of image-grounded generation has therefore become a central goal of recent work on LVLM reliability and hallucination evaluation, spanning training-based fine-tuning, decoding-time guidance, and attention-level interventions.

\subsection{Controlled Generation in LVLMs}
Our work frames hallucination mitigation as a controlled generation problem, aligning with several paradigms for steering generative models at inference time. Contrastive decoding (CD), proposed for language models~\cite{cd, chuang2024doladecodingcontrastinglayers}, steers generation by contrasting logits from a large model with those from a smaller model, and has been adapted to VLMs to mitigate hallucinations~\cite{vcd, lee2024delvevisualcontrastivedecoding, park2025secondmitigatingperceptualhallucination, icd, wan2025onlyonelayerinterventionsufficiently}, for example via visual contrastive decoding (VCD)~\cite{vcd}. 

Similarly, classifier-free guidance (CFG)~\cite{ho2022classifierfreediffusionguidance, sanchez2023staytopicclassifierfreeguidance}, originating in diffusion models, steers generation by combining conditional and unconditional forward passes with a guidance weight to amplify features consistent with the conditioning signal, and this concept has been applied to LVLMs at the logit level~\cite{marine, wan2024contrastiveregionguidanceimproving, zhang2024prompthighlighterinteractivecontrol}. These logit-level methods (visualized in Figure~\ref{fig:main}(a)) are computationally costly, as they require multiple separate forward passes and operate only at the output layer. 

Finally, latent steering (Figure~\ref{fig:main}(b)) intervenes directly on the model's hidden states (representations), an active line of work in LLMs~\cite{subramani-etal-2022-extracting} that has been extended to VLMs~\cite{su-etal-2025-activation, wu-etal-2025-sharp, vti, vista}, typically by adding a precomputed steering vector to the residual stream. 

Our method, ACG, connects but also differs from these lines of work. Unlike CD- or CFG-based methods, which require multiple forward passes, ACG constructs image-conditioned and approximate text-only paths inside the self-attention layers of a single forward pass. Compared to latent steering, which generally utilizes a static steering vector, ACG derives a token-dependent steering direction on the fly from the contrast between image-conditioned and text-only attention outputs, as illustrated in Figure~\ref{fig:main}(d).

\subsection{Attention-based Interventions in LVLMs}
A complementary line of work studies LVLM failures through the lens of the attention mechanism, providing evidence that attention-level biases are closely related to hallucinations and other vision-centric errors~\cite{jiang2025devilsmiddlelayerslarge, adhh, chen2025spatialreasoninghardvlms, zhu2025mitigatingobjecthallucinationslarge}. Prior studies have reported phenomena such as ``text inertia'' (hallucinations persisting even without visual input)~\cite{pai}, fading or noisy visual attention as generation progresses~\cite{jung2025visualattentionfadesselective}, and specialized attention patterns including ``visual attention sinks''~\cite{kang2025toldvisualattentionsink} and ``hallucination heads''~\cite{adhh}. However, current attention-based interventions motivated by these observations (Figure~\ref{fig:main}(c)) remain fragmented and largely heuristic, often targeting specific empirical patterns or relying on additional components such as offline causal analysis to pre-identify heads.

ACG takes a step toward bridging this gap. We define an explicit conditional--unconditional contrast at each attention layer, and introduce an orthogonalized correction that suppresses components aligned with the text-only path. These design choices provide a unified, objective-driven alternative to prior heuristic attention interventions.
\section{Method}
\label{sec:method}
We propose \textbf{Attention-space Contrastive Guidance (ACG)}, a training-free, inference-time guidance mechanism for mitigating hallucinations in large vision--language models (LVLMs). 
This section first introduces the contrastive formulation in attention space, then describes our single-pass approximation of the unconditional path and the associated orthogonalized correction.

\subsection{Preliminaries}
\label{sec:problem_setup}

\noindent\textbf{Architecture.}
We consider LVLMs composed of a vision encoder, a projector, and a LLaMA-style language decoder.  
Given an image $I$ and a text prompt, the vision encoder and projector yield visual embeddings  
$X_v = [x^{(v)}_1, \dots, x^{(v)}_{n_v}]$ in the language embedding space.  
The decoder input is composed of system tokens  
$X_s = [x^{(s)}_1, \dots, x^{(s)}_{n_s}]$, visual tokens $X_v$, user query tokens  
$X_q = [x^{(q)}_1, \dots, x^{(q)}_{n_q}]$, and previously generated response tokens  
$X_{<t} = [x_1, \dots, x_{t-1}]$.  
The complete multimodal context is
\begin{equation}
    X_c = \mathrm{concat}(X_s, X_v, X_q, X_{<t}),
\end{equation}
with total sequence length $n = n_s + n_v + n_q + (t - 1)$.

\medskip
\noindent\textbf{Autoregressive decoding.}
At generation step $t$, the LVLM predicts the next token via
\begin{equation}
    x_t \sim p_\theta(x_t \mid X_c)
    = \mathrm{softmax}\big(H(h^{(L)}_{\text{current}})\big),
\end{equation}
where $h^{(L)}_{\text{current}}$ denotes the last-layer hidden state corresponding to the current decoding position and $H$ is the output projection to logits.
\label{sec:method_algorithm}

\begin{algorithm}[t!]
\caption{Attention-space Contrastive Guidance (ACG)}

% \caption{Attention-space Contrastive Guidance (ACG) in Self-attention Blocks (LLaMA Version)}
\label{alg:acg}
\begin{algorithmic}[1]
\Statex \textbf{Input:} Weights $W_Q, W_K, W_V, W_O$, hidden state $H^{(l-1)}$, and scaling factor $\gamma$
\Statex \textbf{Output:} Updated hidden state $H^{(l)}$
\vspace{0.6em}

\Statex {$\triangleright$ Pre-normalization}
\State $\tilde{H} \gets \mathrm{RMSNorm}(H^{(l-1)})$
\State $Q \gets \tilde{H} W_Q,\;\; K \gets \tilde{H} W_K,\;\; V \gets \tilde{H} W_V$
\vspace{0.6em}

\Statex {$\triangleright$ Conditional and approximate text-only outputs}
\State $S \gets \frac{QK^\top}{\sqrt{d_k}}$
% \State apply standard causal/padding mask
\State $A_{\text{cond}} \gets \mathrm{softmax}(S)$,\;\; $O_{\text{cond}} \gets A_{\text{cond}} V$
\vspace{0.6em}

% \Statex {$\triangleright$ Construct mask $M$}
% \State $M_{ij} = -\infty$ if query $i$ is textual and key $j$ is visual
% \State $A_{\text{uncond}} \gets \mathrm{softmax}(S + M)$,\;\; $O_{\text{uncond}} \gets A_{\text{uncond}} V$
\Statex {$\triangleright$ Construct mask $M$}
\State Let query $i^\star$ be the index of the last text token.
\State $M_{i^\star j} \gets -\infty$ if key $j$ is visual; $M_{ij} \gets 0$ otherwise
\State $A_{\text{uncond}} \gets \mathrm{softmax}(S + M)$,\;\; $O_{\text{uncond}} \gets A_{\text{uncond}} V$
\vspace{0.6em}

\Statex {$\triangleright$ Contrastive correction with orthogonalization}
\State $\Delta O \gets O_{\text{cond}} - O_{\text{uncond}}$
\State $u \gets \frac{O_{\text{uncond}}}{\|O_{\text{uncond}}\|_2 + \epsilon}$
\State $\Delta O_\perp \gets \Delta O - \langle \Delta O, u \rangle u$
\State $O_{\text{final}} \gets O_{\text{cond}} + \gamma \cdot \Delta O_\perp$
\vspace{0.6em}

\Statex{$\triangleright$ Output projection and residual}
\State $Z \gets O_{\text{final}} W_O$
\State $H_{\mathrm{attn}}^{(l)} \gets H^{(l-1)} + Z$
\State $H^{(l)} \gets H_{\mathrm{attn}}^{(l)} + \mathrm{FFN}(\mathrm{RMSNorm}(H_{\mathrm{attn}}^{(l)}))$
\State \textbf{return} $H^{(l)}$
\end{algorithmic}
\end{algorithm}

\subsection{Attention-space Contrastive Guidance}
\label{sec:method_approximation}
Hallucinations often arise when an LVLM’s language priors dominate over its visual evidence.  
Recent work~\cite{adhh} indicates that this failure originates primarily within multi-head attention (MHA) modules rather than MLPs.
Thus, ACG intervenes directly at the source of cross-modal interaction, operating on the \emph{attention output} rather than the final logits as in logit-level contrastive methods~\cite{vcd, sanchez2023staytopicclassifierfreeguidance, pai}.

At each decoding step, we apply ACG to the current response token, i.e., the last text token in the sequence.  
For this query token $q$, we define two attention outputs:
\begin{itemize}
    \item $O_{\text{cond}}$: the \emph{conditional} output, where $q$ attends to all keys.
    \item $O_{\text{uncond}}$: the \emph{text-only (unconditional)} output, representing the model’s behavior without visual conditioning.
\end{itemize}
The guided output is given by the contrastive interpolation
\begin{equation}
\label{eq:base_acg}
    O_{\text{final}} = O_{\text{cond}} + \gamma \cdot (O_{\text{cond}} - O_{\text{uncond}}),
\end{equation}
where $\gamma$ controls the strength of guidance.

However, if we compute $O_{\text{uncond}}$ using a separate forward pass with non-image input, as in prior work~\cite{pai, vista}, this guidance step requires an additional pass and roughly doubles inference time. ACG avoids this cost by \emph{approximating} the unconditional path within a single forward pass using a masking strategy.

Specifically, we compute the query, key, and value matrices ($Q$, $K$, $V$) once per layer.  
The conditional attention output is obtained as:
\begin{equation}
    A_{\text{cond}} = \mathrm{softmax}\!\left(\frac{QK^\top}{\sqrt{d_k}}\right), \quad
    O_{\text{cond}} = A_{\text{cond}} V.
\end{equation}
To approximate the text-only (unconditional) path for the current response token, we reuse the score matrix  
$S = \frac{QK^\top}{\sqrt{d_k}}$ but apply a binary mask $M$ that suppresses attention to visual keys from the last text query. Concretely, for the last text token index $i^\star$,
we set $M_{i^\star j} = -\infty$ if key $j$ is visual and $M_{ij} = 0$ otherwise, and define
\begin{equation}
\label{eq:uncond_approx}
    A_{\text{uncond}} = \mathrm{softmax}(S + M), \quad
    O_{\text{uncond}} = A_{\text{uncond}} V.
\end{equation}
This masking operation effectively removes visual contributions for the current text query, simulating an image-agnostic state while preserving the same computational graph and reusing all intermediate states. We expect this single-pass approximation to capture the language-biased behavior that underlies hallucination.

\subsection{Textual Orthogonalization}
\label{sec:method_ortho}

While efficient, the masking-based approximation introduces an inherent \emph{approximation bias}: the masked $O_{\text{uncond}}$ does not perfectly match a true image-absent forward pass. We highlight two main sources of this mismatch:
\begin{enumerate}
    \item \textbf{Contextual leakage.}
    Because earlier layers may already inject visual context into $Q$, $K_{\text{text}}$, and $V_{\text{text}}$, masking visual keys at layer $l$ does not in general reproduce a truly image-absent state at that layer.
    \item \textbf{Softmax redistribution.}
    When visual keys are masked, attention mass that would otherwise target visual tokens is redistributed to text tokens, amplifying text--text correlations and altering the effective language prior.
\end{enumerate}

As a result, the naive guidance vector $\Delta O = O_{\text{cond}} - O_{\text{uncond}}$ can mix the desired visual correction with text-induced distortion, which may degrade response quality at higher guidance scales $\gamma$.

To mitigate this effect, ACG applies \textbf{textual orthogonalization}, a geometric correction that suppresses components of the guidance signal aligned with the text-only path. We treat $O_{\text{uncond}}$ as defining a principal textual direction and remove from $\Delta O$ the component parallel to that direction.

First, we define the normalized direction
\begin{equation}
    u = \frac{O_{\text{uncond}}}{\|O_{\text{uncond}}\|_2 + \epsilon},
\end{equation}
where $\epsilon$ ensures numerical stability. We then project $\Delta O$ onto the subspace orthogonal to $u$:
\begin{equation}
\label{eq:orthogonalization}
    \Delta O_{\perp} = \Delta O - \langle \Delta O, u \rangle u.
\end{equation}
This operation removes the component of $\Delta O$ parallel to the text-only path, yielding a guidance update that is less influenced by the text-only direction. The final guided output is
\begin{equation}
\label{eq:final_acg}
    O_{\text{final}} = O_{\text{cond}} + \gamma \cdot \Delta O_{\perp}.
\end{equation}
This correction strengthens the contrastive update while reducing drift along the textual direction, improving the stability of guidance at higher $\gamma$. We apply this correction at every decoding step across all self-attention layers used by ACG, and summarize the full procedure in Algorithm~\ref{alg:acg}.

\section{Experiments}
\label{sec:exp}
\subsection{Experimental Settings}
\begin{table*}[t]
\centering
\caption{
\textbf{POPE results (\%).}
We report overall average accuracy (Avg.) as well as accuracy (Acc.), precision (Prec.), recall (Rec.), and F1 under the \textit{Random}, \textit{Popular}, and \textit{Adversarial} settings.
\textbf{Bold} indicate the best performance in each column.
}
\label{tab:pope_main}
\setlength{\tabcolsep}{4pt}
% \scriptsize
\resizebox{0.9\textwidth}{!}{
\begin{tabular}{l l c cccc cccc cccc}
\toprule
\multirow{2}{*}{Model}&\multirow{2}{*}{Method}& \multicolumn{1}{c}{Avg.} &
\multicolumn{4}{c}{Random} &
\multicolumn{4}{c}{Popular} &
\multicolumn{4}{c}{Adversarial} \\
\cmidrule(lr){3-3}
\cmidrule(lr){4-7}
\cmidrule(lr){8-11}
\cmidrule(lr){12-15}
 &  &
Acc. &
Acc. & Prec. & Rec. & F1 &
Acc. & Prec. & Rec. & F1 &
Acc. & Prec. & Rec. & F1 \\
\midrule\midrule
\multirow{5}{*}{LLaVA~\cite{llava1_5}}
  & Regular & 84.83 & \textbf{89.37} & 89.66 & 89.00 & \textbf{89.33} &
                         86.00 & 84.05 & 88.87 & 86.39 &
                         79.13 & 74.39 & 88.87 & 80.98 \\
  & VCD~\cite{vcd}     & 85.38 & 88.63 & 91.96 & 84.67 & 88.16 &
                         85.97 & 86.93 & 84.67 & 85.78 &
                         81.53 & 79.67 & 84.67 & 82.09 \\
  & PAI~\cite{pai}     & 84.91 & 89.30 & 89.54 & 89.00 & 89.27 &
                         86.13 & 84.26 & 88.87 & \textbf{86.50} &
                         79.30 & 74.59 & 88.87 & 81.11 \\
  & VISTA~\cite{vista}   & 83.03 & 88.25 & 87.31 & \textbf{90.33} & 88.79 &
                         84.10 & 80.21 & \textbf{90.53} & 85.06 &
                         76.73 & 70.93 & \textbf{90.60} & 79.57 \\
\rowcolor{blue!20}\cellcolor{white}
  & Ours    & \textbf{86.03} & 88.30 & \textbf{95.92} & 80.00 & 87.24 &
                        \textbf{ 86.57} & \textbf{92.22} & 79.87 & 85.60 &
                         \textbf{83.23} & \textbf{85.63} & 79.87 & \textbf{82.65} \\
\midrule
\multirow{4}{*}{MiniGPT-4~\cite{zhu2023minigpt4enhancingvisionlanguageunderstanding}}
  & Regular & 76.31 & 82.47 & \textbf{89.27 }& 73.80 & 80.80 &
                         75.00 & 75.48 & 74.07 & 74.76 &
                         71.47 & 70.41 & 74.07 & 72.19 \\
  & PAI     & 76.59 & 82.23 & 88.53 & 74.07 & 80.65 &
                         \textbf{75.80} & \textbf{76.62} & 74.27 & 75.42 &
                         71.73 & \textbf{70.69} & 74.27 & 72.43 \\
  & VISTA   & 76.19 & 82.30 & 87.40 &\textbf{ 76.73 }& \textbf{81.72} &
                         75.10 & 74.18 & \textbf{77.00} & \textbf{75.56} &
                         71.17 & 69.07 & \textbf{76.67} & 72.67 \\
\rowcolor{blue!20}\cellcolor{white}
  & Ours    & \textbf{76.70 }& \textbf{82.70} & 88.90 & 74.73 & 81.20 &
                         75.50 & 75.76 & 75.00 & 75.38 &
                         \textbf{71.90} & 70.62 & 75.00 & \textbf{72.74} \\
\midrule
\multirow{3}{*}{Qwen-VL~\cite{bai2023qwenvlversatilevisionlanguagemodel}}
  & Regular & 85.51 & 86.67 & \textbf{98.50} & 74.47 & 84.81 &
                         85.80 & \textbf{96.29} & 74.47 & 83.98 &
                         84.07 & \textbf{92.23} & 74.40 & 82.36 \\
  & VCD     & 86.77 & 88.83 & 94.64 & 82.33 & 88.06 &
                         87.23 & 90.98 & 82.67 & 86.62 &
                         \textbf{84.23} & 85.44 & 82.53 & \textbf{83.96 }\\
\rowcolor{blue!20}\cellcolor{white}
  & Ours    & \textbf{86.98} & \textbf{89.17} & 93.81 & \textbf{83.87} & \textbf{88.56} &
                         \textbf{88.23} & 91.89\textbf{ }& \textbf{83.87} & \textbf{87.70} &
                         83.53 & 83.27 & \textbf{83.93} & 83.60 \\
\bottomrule
\end{tabular}
}
\normalsize
\end{table*}

% \noindent\textbf{Datasets and Benchmarks.}
% \noindent\textbf{Benchmarks.}

%We evaluate on two widely used hallucination benchmarks, POPE~\cite{pope} and CHAIR~\cite{chair}, built on %MS~COCO. For further evaluation, we choose MMHal-Bench~\cite{mmhal}.
%POPE measures binary yes/no object existence, while CHAIR evaluates object hallucinations in free-form captions.
%MMHal-Bench consists of 96 image–question pairs that probe object and attribute-level inconsistencies. Model %responses' alignment with ground-truth answers is evaluated by GPT-4.
We evaluate our method on both hallucination benchmarks and general multimodal benchmarks.
For hallucination evaluation, we consider object-level and attribute-level settings.
Specifically, we use POPE~\cite{pope} and CHAIR~\cite{chair}, both built on MS COCO, for object-level hallucination, and MMHal-Bench~\cite{mmhal} for attribute-level hallucination.
POPE measures binary yes/no object existence, while CHAIR evaluates object hallucinations in free-form captions.
MMHal-Bench consists of 96 image–question pairs probing both object- and attribute-level inconsistencies, where the alignment between model responses and ground-truth answers is evaluated using GPT-4.
For broader evaluation beyond hallucination benchmarks, we additionally use MMMU~\cite{yue2023mmmu} and MathVista~\cite{lu2024mathvistaevaluatingmathematicalreasoning}.
MMMU assesses broad multimodal knowledge and complex visual reasoning, and we follow the standard protocol to report accuracy based on exact match or multiple-choice selection.
MathVista focuses on visual mathematical reasoning, and we report answer accuracy by comparing model predictions with ground-truth answers, applying appropriate normalization for numerical responses. 

\noindent\textbf{Baselines.}
We compare ACG against three representative inference-time approaches:
(i) logit-level contrastive decoding (VCD~\cite{vcd}),
(ii) logit-level classifier-free guidance with attention intervention (PAI~\cite{pai}),
and (iii) latent steering (VISTA~\cite{vista}).
These baselines cover output-logit, latent-space, and attention-space interventions for LVLMs. For each baseline, we evaluate only on LVLMs officially supported by the authors' public implementations, to avoid unfair or unstable re-implementations.

\noindent\textbf{Models and Implementation Details.}
We consider three LVLMs with diverse language backbones and vision-language connectors:
LLaVA-1.5~\cite{llava1_5},
MiniGPT-4~\cite{zhu2023minigpt4enhancingvisionlanguageunderstanding},
and Qwen-VL~\cite{bai2023qwenvlversatilevisionlanguagemodel}. ACG exposes a single tunable hyperparameter, the guidance scale $\gamma$. Unless otherwise stated, we use $\gamma=2.4$ for LLaVA-1.5, $\gamma=0.3$ for MiniGPT-4, and $\gamma=1.4$ for Qwen-VL. The model-specific values reflect architectural differences. For MMHal, MMMU, and MathVista on LLaVA-1.5, we use a slightly smaller value, $\gamma=2.0$, for more stable behavior. We use greedy decoding throughout for consistent comparison across methods and benchmarks.

\subsection{Results}
\noindent\textbf{Results on POPE.}
Results on POPE under the random, popular, and adversarial settings, together with the overall average, are shown in \cref{tab:pope_main}. ACG outperforms the baselines in overall average score. The improvement is particularly notable on the adversarial split for LLaVA-1.5 and MiniGPT-4, where negative samples are semantically or statistically related to objects that do appear in the image. This result suggests that ACG more effectively suppresses language-biased object predictions under challenging confusable settings.

\begin{table}[t]
\centering
\caption{
\textbf{CHAIR results (\%) on LLaVA-1.5 and MiniGPT-4.}
\textbf{Bold} indicates the best CHAIR performance and \underline{underlines} indicate the second-best in each column.
}\label{tab:chair_main}
\setlength{\tabcolsep}{3pt}
% \scriptsize
\resizebox{\columnwidth}{!}{%
\begin{tabular}{l l ccc ccc}
\toprule
\multirow{2}{*}{Model}&\multirow{2}{*}{Method}& \multicolumn{3}{c}{Max Tokens 128} & \multicolumn{3}{c}{Max Tokens 64} \\
\cmidrule(lr){3-5} \cmidrule(lr){6-8}
 &  & CHAIR$_s$ & CHAIR$_i$ & F1 & CHAIR$_s$ & CHAIR$_i$ & F1 \\
\midrule
\midrule
\multirow{5}{*}{LLaVA~\cite{llava1_5}}
  & Regular & 56.2 & 18.3 & 70.6 & 25.2 &  8.3 & 68.5 \\
  & VCD     & 55.0 & 17.0 & 72.5 & 27.2 &  8.8 & 70.0 \\
  & PAI     & \underline{25.6} &  \underline{7.6} & 75.9 & \textbf{13.8} &  \textbf{4.5} & 72.4 \\
  & VISTA   & 31.0 & 10.5 & 76.6 & 23.6 &  7.6 & 74.3 \\
\rowcolor{blue!20}\cellcolor{white}
  & Ours    & \textbf{21} &  \textbf{4.8} & 74.4 & \underline{16.8} &  \textbf{4.5}& 72.8 \\
\midrule
\multirow{4}{*}{MiniGPT-4~\cite{zhu2023minigpt4enhancingvisionlanguageunderstanding}}
  & Regular & 35.0 & 10.8 & 69.8 & 24.8 &  8.2 & 69.3 \\
  & PAI     & 22.8 &  8.1 & 71.4 & 19.8 &  6.9 & 71.0 \\
  & VISTA   & \underline{18.8} &  \underline{5.9} & 71.0 & \underline{15.8} &  \underline{5.5} & 70.2 \\
\rowcolor{blue!20}\cellcolor{white}
  & Ours    & \textbf{10.8} &  \textbf{3.3} & 68.0 &\textbf{ 11.2 }& \textbf{4.2} & 67.9 \\
\bottomrule
\end{tabular}%
}
\normalsize
\end{table}

\noindent\textbf{Results on CHAIR.} 
Results on open-ended generation with CHAIR are shown in \cref{tab:chair_main}. We report CHAIR$_s$, CHAIR$_i$, and F1 under $\texttt{max\_new\_tokens}\in\{64,128\}$ to disentangle the effect of caption length from hallucination reduction.

Across both models and both length budgets, ACG consistently achieves the lowest CHAIR$_i$. On LLaVA-1.5, it reduces CHAIR$_i$ to 4.8 and CHAIR$_s$ to 21.0 at 128 tokens while keeping F1 close to the strongest baseline. On MiniGPT-4, it attains the best CHAIR$_s$ and CHAIR$_i$ in both settings with only a mild drop in F1. Overall, ACG substantially reduces hallucinations while largely preserving object-level fidelity, and its gains remain strong under longer generation.

% Results on MMMU, MMHal, and MathVista with LLaVA-1.5.}
\begin{table}[t]
\centering
\caption{\textbf{Performance on MMHal, MMMU, and MathVista.}}
% Our method improves over the baseline on MMMU, MMHal, and MathVista.}
\label{tab:mmmu_mmhal_mathvista}
\small
\setlength{\tabcolsep}{3pt}
\begin{tabular}{lccccc}
\toprule
 & MMHal (\(\uparrow\)) & \multicolumn{3}{c}{MMMU (\(\uparrow\))} & MathVista (\(\uparrow\)) \\
\cmidrule(lr){3-5}
Method &  & MC & Open & Total &  \\
\midrule
Regular & 1.94 & 37.54 & 3.77 & 35.56 & 22.6 \\
%\rowcolor{blue!10}
\rowcolor{blue!20}
Ours ($\gamma$=2) & \textbf{2.12} & \textbf{38.49} & \textbf{9.43} & \textbf{36.78} & \textbf{23.7} \\
\bottomrule
\end{tabular}
\vspace{-0.4cm}
\end{table}

\noindent\textbf{Results on MMHal-Bench.}
We evaluate the effect of our method on attribute-level hallucination using MMHal-Bench.
As shown in \cref{tab:mmmu_mmhal_mathvista}, our method improves over the baseline on the LLaVA-1.5 architecture.
These results demonstrate that our approach enhances robustness to both object- and attribute-level hallucination. Detailed category-wise MMHal-Bench results and a sweep over multiple $\gamma$ values are provided in the supplement.
% Note that, For this evaluation, we use $\gamma = 2.0$, as a slightly weaker weighting yields more stable behavior.

%To assess the effect of our proposed method on logical reasoning and complex visual understanding, we conduct %comparison between the baseline and our model on the LLaVA-1.5 architecture, as illustrated in %\cref{tab:mmmu_mmhal_mathvista}. 
%%The radar chart reports hallucination-related metrics across eight categories: %object attributes (ATTR), %adversarial objects (ADV), comparisons (COMP), counting %(COUNT), spatial relations (SPAT), environmental %inferences (ENV), holistic %descriptions (HOL), and others (OTHER), where higher scores indicate better %factual %alignment and reduced hallucination.
%Overall, our method consistently outperforms the baseline on average score. It confirms that our approach enhances %robustness against both object-level and attribute-level hallucination. 
%Note that, for this evaluation, we adopt $\gamma = 2.0$, as a slightly weaker weighting yielded more stable %behavior in this controlled comparison.

\noindent\textbf{Generalization Beyond Hallucination Benchmarks.}
To evaluate whether our method generalizes beyond hallucination benchmarks, we assess its performance on two general multimodal benchmarks: MMMU~\cite{yue2023mmmu} and MathVista~\cite{lu2024mathvistaevaluatingmathematicalreasoning}.
% As shown in \cref{tab:mmmu_mmhal_mathvista}, ACG slightly improves over the baseline on both benchmarks. 
As shown in \cref{tab:mmmu_mmhal_mathvista}, our method achieves slightly improved performance over the baseline on MMMU, which evaluates broad multimodal knowledge and complex visual reasoning. 
% This suggests that our method improves general reasoning performance while reducing hallucinations. 
On MathVista, which focuses on visual mathematical reasoning, our method similarly shows consistent improvements in answer accuracy compared to the baseline. 
% These results indicate that our approach preserves—and can slightly enhance—numerical reasoning capabilities. 
Overall, the results demonstrate that our method effectively mitigates hallucinations while maintaining and slightly improving performance on general multimodal tasks.

\subsection{Generalization to Newer and Larger LVLMs}
\label{sec:newer_larger_models}

To examine whether ACG generalizes beyond the original model set, we additionally evaluate it on LLaVA-NeXT 7B and 13B using the same protocol as in our main experiments, and test CHAIR under a longer decoding budget with \texttt{max\_new\_tokens=512}.

As shown in Table~\ref{tab:chair_llavanext}, ACG with a moderate guidance strength ($\gamma=2.0$) consistently reduces hallucination metrics on both models while maintaining or improving F1. On LLaVA-NeXT 7B, ACG reduces CHAIR$_s$ from 31.2 to 25.2 and CHAIR$_i$ from 8.1 to 5.4, while improving F1 from 72.1 to 73.9. On LLaVA-NeXT 13B, it reduces CHAIR$_i$ from 8.3 to 5.5 and improves F1 from 71.5 to 74.9, while also lowering CHAIR$_s$ from 33.8 to 31.0. These results show that ACG remains effective on newer and larger LVLMs, and under longer decoding budgets. 
% A full guidance sweep is provided in the supplement.

\begin{table}[t]
\centering
\caption{
\textbf{Generalization results (\%) on LLaVA-NeXT 7B and 13B under longer decoding.}
We evaluate CHAIR with \texttt{max\_new\_tokens=512}. We use $\gamma=2.0$ for both models. 
% \textbf{Bold} indicates the better CHAIR performance in each model.
}
\label{tab:chair_llavanext}
\setlength{\tabcolsep}{4pt}
\renewcommand{\arraystretch}{1.1}
\resizebox{0.9\columnwidth}{!}{%
\begin{tabular}{l l ccc}
\toprule
Model & Method & CHAIR$_s$ & CHAIR$_i$ & F1 \\
\midrule
LLaVA-NeXT 7B  & Regular & 31.2 & 8.1 & 72.1 \\
\rowcolor{blue!20}
LLaVA-NeXT 7B  & Ours & \textbf{25.2} & \textbf{5.4} & \textbf{73.9} \\
\midrule
LLaVA-NeXT 13B & Regular & 33.8 & 8.3 & 71.5 \\
\rowcolor{blue!20}
LLaVA-NeXT 13B & Ours & \textbf{31.0} & \textbf{5.5} & \textbf{74.9} \\
\bottomrule
\end{tabular}%
}
\renewcommand{\arraystretch}{1.0}
\normalsize
\end{table}
\section{Analysis}
\label{sec:analysis}
% We analyze the validity of the masked surrogate, the effect of textual orthogonalization, the sensitivity to guidance strength, and the efficiency of ACG.

\subsection{Justifying the Masked Unconditional Path}
\label{sec:masked_uncond}

\begin{figure}[t]
\centering
\includegraphics[width=0.90\columnwidth]{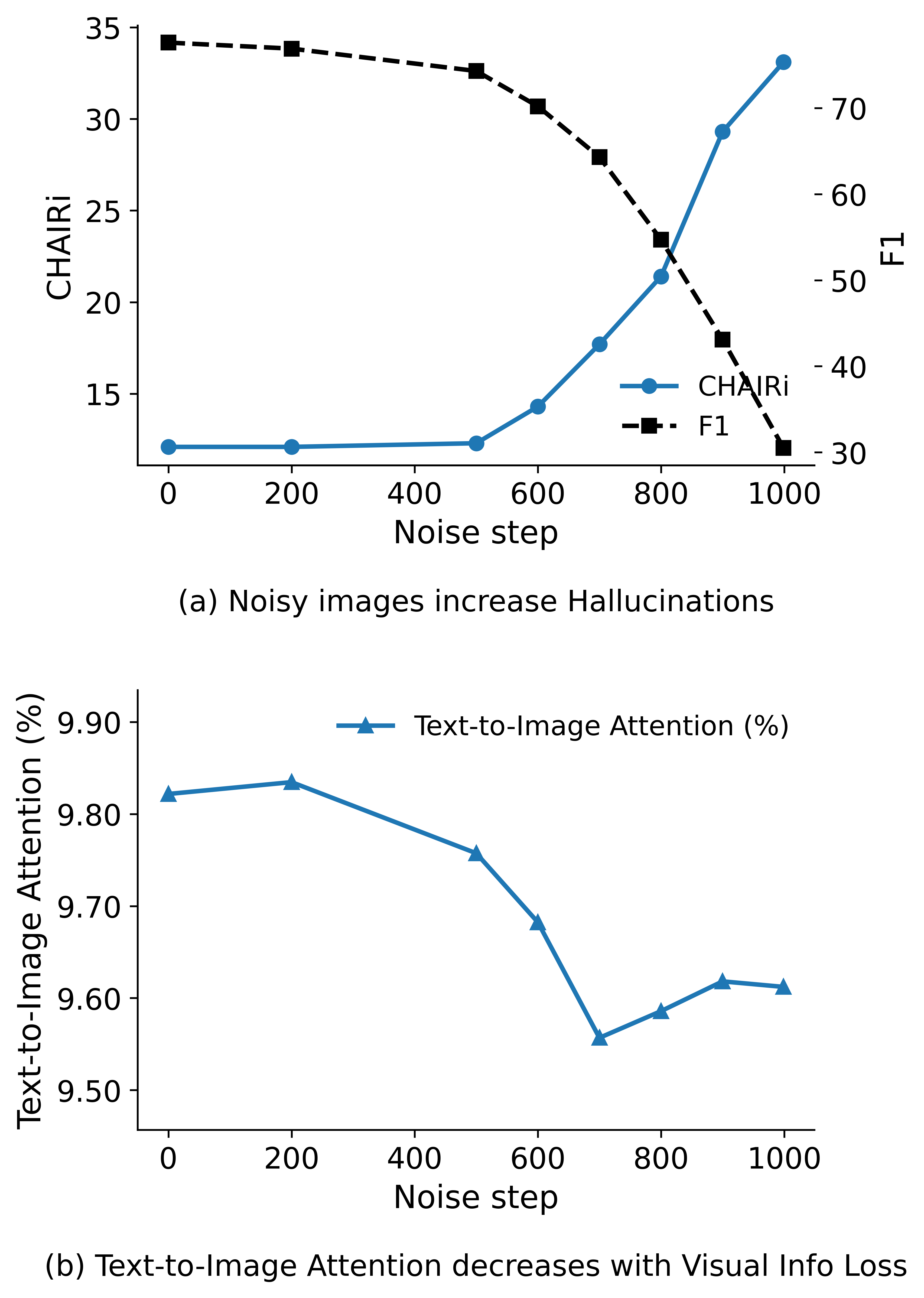}
\caption{\textbf{Justifying the masked unconditional path.}
As visual information is degraded by Gaussian noise, hallucination increases, fidelity decreases, and the mean text-to-image (T2I) attention ratio shows an overall downward trend.}
\label{fig:noise_justification}
\end{figure}

ACG relies on a single-pass surrogate of the text-only path, $O_{\text{uncond}}^{\text{mask}}$, obtained by masking visual keys in attention (Sec.~\ref{sec:method}). Here, we provide empirical evidence that this surrogate tracks the ungrounded, language-prior regime that emerges when visual evidence becomes weak or uninformative.

\noindent\textbf{Protocol.}
We progressively degrade the input image by adding Gaussian noise (noise step $\in \{0,\ldots,999\}$), run the vanilla LLaVA model without guidance, and measure
(i) instance-level hallucination (CHAIR$_i$),
(ii) object-level fidelity (F1), and
(iii) the mean text-to-image (T2I) attention ratio, averaged across all generated tokens, layers, and heads.

\noindent\textbf{Finding 1: Visual Information Loss Correlates with Hallucination.}
As shown in Fig.~\ref{fig:noise_justification}(a), increasing noise is associated with a severe loss of faithfulness: CHAIR$_i$ rises sharply from 12.1 to 33.1, while F1 drops from 77.6 to 30.5, with a pronounced knee near the 600-step mark.
Consistent with prior observations that visual corruption exacerbates hallucination~\cite{vti}, this trend supports our hypothesis that removing visual evidence pushes the model toward a more text-biased, high-hallucination regime. 
This in turn motivates the text-only regime as a meaningful reference point for intervention.

\noindent\textbf{Finding 2: The Model Naturally Downweights Uninformative Inputs.}
Fig.~\ref{fig:noise_justification}(b) provides complementary mechanistic evidence. As noise is added, the model's average T2I attention shows an overall downward trend, decreasing from 9.82\% to a minimum of 9.56\% at the 700-step mark. This suggests that, when visual inputs become less informative, the model naturally reduces visual grounding.

% We observe a minor rebound in T2I attention after this 700-step tipping point. However, this non-monotonicity does not change the main picture: over the same region (700~$\to$~999 steps), faithfulness continues to deteriorate rapidly (Fig.~\ref{fig:noise_justification}(a)), indicating that the model remains in an ungrounded, high-hallucination regime despite the small attention rebound.

\noindent\textbf{Summary.}
Taken together, degrading visual inputs induces
(i) rising hallucinations and collapsing fidelity, and
(ii) an overall reduction in T2I attention.
These findings support our use of $O_{\text{uncond}}^{\text{mask}}$ as a principled, single-pass proxy for the model's response when visual evidence becomes weak, while direct comparisons to a true text-only trajectory are provided in the supplement.

\begin{table}[t]
\centering
\caption{\textbf{Ablation at matched F1.} At similar object-level fidelity (F1), ACG (w/ Ortho) yields lower CHAIR$_s$ and CHAIR$_i$ than ACG (w/o Ortho).}
\label{tab:ortho_ablation}
\setlength{\tabcolsep}{3pt}
\scriptsize
\resizebox{0.8\columnwidth}{!}{%
\begin{tabular}{l c c c c}
\toprule
Method & $\gamma$ & F1 $\uparrow$ & CHAIR$_s$ $\downarrow$ & CHAIR$_i$ $\downarrow$ \\
\midrule\midrule
\rowcolor{blue!20}
ACG (w/ Ortho)   & 2.1 & 77.6 & \textbf{34.2} & \textbf{7.6} \\
ACG (w/o Ortho)  & 1.2 & 77.4          & 38.8          & 9.7 \\
\midrule
\rowcolor{blue!20}
ACG (w/ Ortho)   & 2.4 & 74.4 & \textbf{21.0} & \textbf{4.8} \\
ACG (w/o Ortho)  & 1.3 & 74.0          & 30.4          & 8.8 \\
\bottomrule
\end{tabular}%
}
\normalsize
\end{table}

\subsection{Effect of Textual Orthogonalization}
Our primary component, textual orthogonalization, is designed to mitigate the approximation bias introduced by the masking-based construction of $O_{\text{uncond}}^{\text{mask}}$ that enables our single-pass algorithm. We hypothesize that this bias contaminates the naive guidance vector $\Delta O$, so that reducing hallucinations comes at an unnecessarily large cost in object-level fidelity (CHAIR F1).
To test this, we conduct a controlled ablation comparing \textbf{ACG with Ortho} against a naive \textbf{ACG without Ortho}. We choose guidance scales that yield similar F1, and then compare sentence-level and instance-level faithfulness (CHAIR$_s$, CHAIR$_i$).

Table~\ref{tab:ortho_ablation} provides clear evidence for the benefit of orthogonalization. At the $\approx 74$ F1 operating point, ACG (w/ Ortho) attains 1.8$\times$ lower CHAIR$_i$ (8.8 $\to$ 4.8) and 1.4$\times$ lower CHAIR$_s$ (30.4 $\to$ 21.0) than ACG (w/o Ortho). Thus, suppressing the text-aligned component of $\Delta O$ substantially reduces hallucinations while keeping F1 nearly unchanged. This supports orthogonalization as an effective correction for the bias introduced by the masked surrogate. Additional analysis in the supplement further shows that this correction better aligns the resulting guidance with the intended contrastive direction.

% \subsection{Effect of Textual Orthogonalization}
% Our primary component, textual orthogonalization, is designed to correct the approximation bias introduced by the masking-based construction of $O_{\text{uncond}}^{\text{mask}}$ that enables our single-pass algorithm. We hypothesize that this bias contaminates the naive guidance vector $\Delta O$, so that reducing hallucinations comes at an unnecessarily large cost in object-level fidelity (CHAIR F1).
% To test this, we run a controlled ablation comparing \textbf{ACG with Ortho} against a naive \textbf{ACG without Ortho}. We choose guidance scales that yield similar F1 and then compare sentence-level and instance-level faithfulness (CHAIR$_s$, CHAIR$_i$).

% Table~\ref{tab:ortho_ablation} provides strong evidence for the benefit of orthogonalization. At the $\approx 74$ F1 operating point, ACG (w/ Ortho) attains 1.8$\times$ lower CHAIR$_i$ (8.8 $\to$ 4.8) and 1.4$\times$ lower CHAIR$_s$ (30.4 $\to$ 21.0) than ACG (w/o Ortho). Thus, removing the text-aligned component of $\Delta O$ allows us to substantially reduce hallucinations while keeping F1 nearly unchanged, leading to state-of-the-art faithfulness among evaluated inference-time methods.

\subsection{Guidance Scale and Layer-wise Analysis}
\label{sec:param_analysis}

\begin{figure}[t]
  \centering
  \includegraphics[width=0.95\columnwidth]{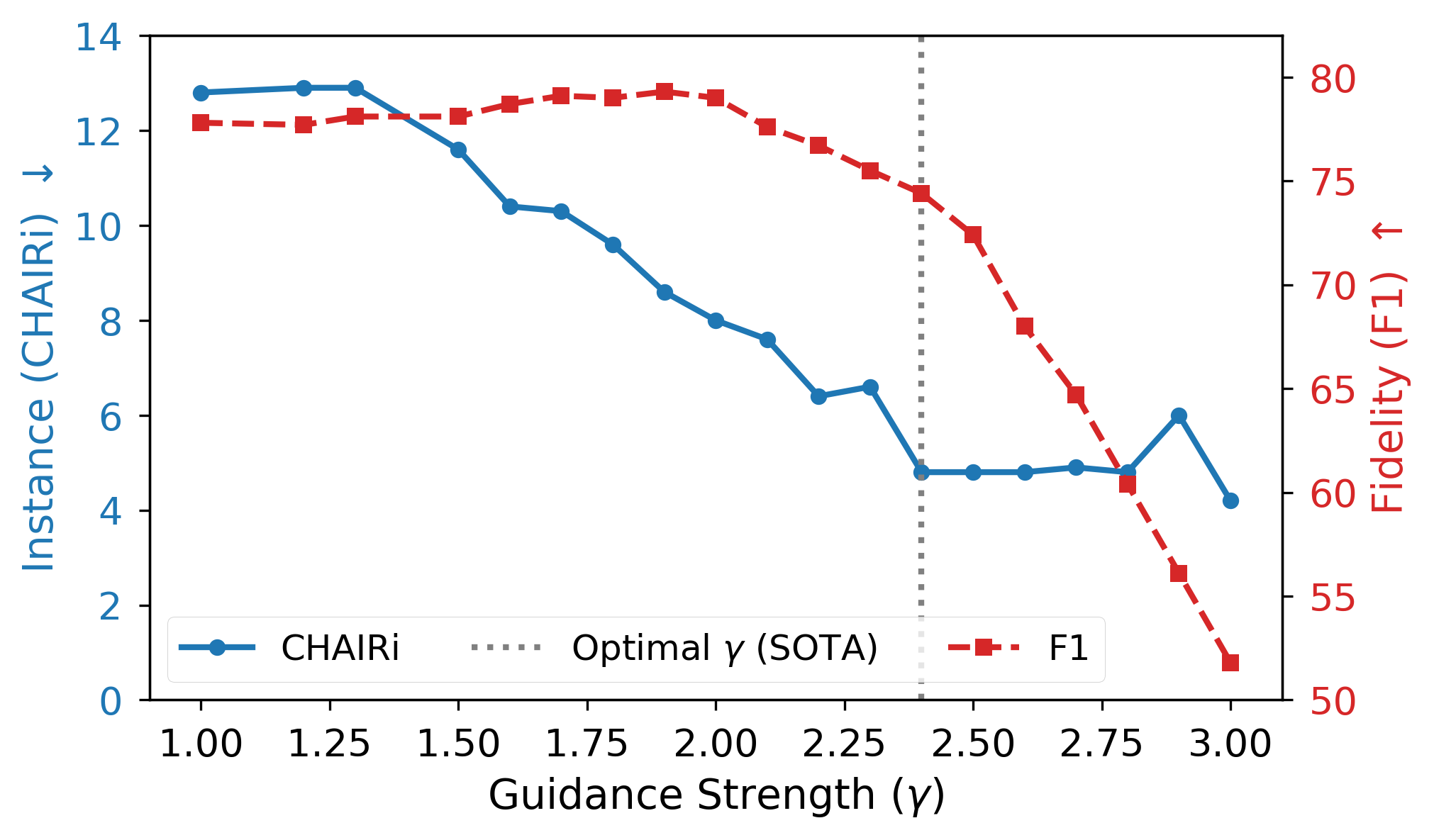}
  % \caption{\textbf{Guidance–scale trade-off on LLaVA-1.5 (CHAIR, max 128).}
  % Increasing $\gamma$ lowers instance hallucination (CHAIR$_i$, blue; left axis) but eventually reduces object-level fidelity (F1, red; right axis). The dotted line marks the canonical setting $\gamma{=}2.4$.}
  \caption{\textbf{Guidance-scale trade-off on LLaVA-1.5 (CHAIR, max 128).}
Increasing $\gamma$ reduces instance-level hallucination (blue; left axis), but overly large $\gamma$ degrades object-level fidelity (red; right axis). The dotted line marks the canonical setting $\gamma{=}2.4$.}

  \label{fig:gamma_tradeoff}
\end{figure}

% ACG exposes a single main hyperparameter, the guidance scale $\gamma$, and by default applies guidance to all layers, without requiring layer-selection tuning. In this subsection, we analyze how $\gamma$ governs the trade-off among faithfulness, fidelity, and caption length, and then examine where along the decoder stack guidance is most effective.

\noindent\textbf{Guidance scale trade-off.}
Figure~\ref{fig:gamma_tradeoff} summarizes a sweep over $\gamma\!\in[1.0,3.0]$ on CHAIR (max 128).
As $\gamma$ increases, instance hallucination (CHAIR$_i$) decreases from $12.8$ at $\gamma{=}1.0$ to around $5$ near $\gamma{=}2.4$, while object-level fidelity (F1) stays in the high 70s up to this range.
Beyond $\gamma{\approx}2.4$, F1 drops sharply and captions become overly short.
We therefore adopt $\gamma{=}2.4$ as a canonical operating point, which achieves strong hallucination reduction (CHAIR$_i{=}4.8$) while maintaining acceptable fidelity (F1$=74.4$) and reasonable caption length.
% \noindent\textbf{Guidance scale trade-off.} 
% Figure~\ref{fig:gamma_tradeoff} summarizes a sweep over $\gamma\!\in[1.0,3.0]$ on CHAIR (max 128). As $\gamma$ increases, instance hallucination (CHAIR$_i$) decreases overall, from $12.8$ at $\gamma{=}1.0$ to $4.8$ at $\gamma{=}2.4$ and $4.2$ at $\gamma{=}3.0$ . Object-level fidelity (F1) remains roughly stable in the $77{\sim}79$ range up to a knee around $\gamma\!\approx\!2.1$–$2.4$, after which it collapses ($74.4$ at $\gamma{=}2.4$, $51.8$ at $\gamma{=}3.0$). Caption length shortens in parallel ($91.6 \rightarrow 72.4 \rightarrow 25.8$). We therefore report a canonical operating point at $\gamma{=}2.4$, which attains strong faithfulness (CHAIR$_i{=}4.8$) while preserving acceptable fidelity (F1$=74.4$).

\begin{table}[t]
\centering
\caption{\textbf{Block-wise characterization on LLaVA-1.5 (CHAIR, max 128).} We partition the decoder into contiguous layer blocks and report representative operating points from each $\gamma$ sweep.}
\label{tab:blockwise_ablation}
\scriptsize
\setlength{\tabcolsep}{5pt}
\resizebox{0.9\columnwidth}{!}{
\begin{tabular}{lcccc}
\toprule
Layer Block & $\gamma$ & CHAIR$_i$ (↓) & F1 (↑) & Len (↓) \\
\midrule\midrule
\textbf{All (1–32)} & \textbf{2.4} & \textbf{4.8} & 74.4 & 72.4 \\
\midrule
Early (1–8) & 2.5 & 7.1 & 77.5 & 77.6 \\
Early (1–8) & 3.0 & 5.3 & 69.8 & 68.0 \\
\midrule
Early–Mid (9–16) & 6.0 & 11.1 & 74.8 & 79.2 \\
Early–Mid (9–16) & 8.0 & 5.7 & 56.6 & 45.9 \\
\midrule
Mid–Late (17–24) & 6.0 & 10.7 & 77.4 & 95.6 \\
Mid–Late (17–24) & 10.0 & 7.0 & 73.4 & 91.8 \\
\midrule
Late (25–32) & 2.5 & 10.7 & 78.6 & 93.0 \\
Late (25–32) & 10.0 & 8.8 & 75.7 & 90.7 \\
\bottomrule
\end{tabular}
}
\end{table}

% \begin{table}[t]
% \centering
% \caption{Efficiency vs.\ faithfulness on LLaVA-1.5 (CHAIR, max 128).
% All values are averages; single-pass methods require one forward pass per generation. Lower is better for latency and CHAIR$_i$.}
% \label{tab:efficiency}
% \scriptsize
% \setlength{\tabcolsep}{5pt}
% \begin{tabular}{lccccc}
% \toprule
% Method & Level & Passes & Avg.\ Latency (s) $\downarrow$ & CHAIR$_i$ $\downarrow$ \\
% \midrule\midrule
% Regular    & --      & 1-pass & 2.81 (1.00$\times$) & 18.3 \\
% VCD          & Logit   & 2-pass & 5.54 (1.97$\times$) & 17.0 \\
% PAI     & Logit+Attn & 2-pass & 6.42 (2.28$\times$) & 7.6 \\
% VISTA       & Latent  & 3-pass & -($\times$) & 10.5 \\
% \midrule
% ACG-Fast  & Attention & \textbf{1-pass} & \textbf{2.96 (1.05$\times$)} & 7.3 \\
% ACG-Full & Attention   & \textbf{1-pass} & 3.34 (1.19$\times$) & \textbf{4.8} \\
% \bottomrule
% \end{tabular}
% \end{table}

% Preamble에 이 라인을 추가해야 합니다: \usepackage{makecell}

\begin{table}[t]
\centering
\caption{\textbf{Efficiency vs.\ faithfulness on LLaVA-1.5 (CHAIR, max 128).} We compare latency, number of forward passes, and CHAIR$_i$. All values are averaged over the evaluation set.}
\label{tab:efficiency}
\scriptsize
\setlength{\tabcolsep}{5pt}
\begin{tabular}{lcccccc} 
\toprule
Method & Level & Passes & \makecell{Per-image \\ Latency (s) $\downarrow$} & \makecell{Per-word \\ Latency (s) $\downarrow$} & \makecell{CHAIR$_i$ \\ $\downarrow$} \\
\midrule\midrule
Regular & -- & 1-pass & 2.81 (1.00$\times$) & 0.03 & 18.3 \\
\midrule
VCD & Logit & 2-pass & 5.54 (1.97$\times$) & 0.06 & 17.0 \\
PAI & Logit+Attn & 2-pass & 6.42 (2.28$\times$) & 0.07 & 7.6 \\
VISTA & Latent & 3-pass & 5.55 (1.98$\times$) & 0.07 & 10.5 \\
\midrule
ACG-Fast & Attention & \textbf{1-pass} & \textbf{2.96 (1.05$\times$)} & \textbf{0.04} & 7.3 \\
ACG-Full & Attention & \textbf{1-pass} & 3.34 (1.19$\times$) & 0.05 & \textbf{4.8} \\
\bottomrule
\end{tabular}
\end{table}

\noindent\textbf{Block-wise characterization.}
While our default setup uses \emph{All} layers, we further ask where guidance is most effective by partitioning the 32-layer decoder of LLaVA-1.5 into four contiguous blocks and sweeping $\gamma$ within each block. 

Table~\ref{tab:blockwise_ablation} reports representative operating points along each curve. Applying guidance to \emph{Early} layers already yields substantial hallucination reduction at modest scales, whereas \emph{All} layers achieve the strongest reduction overall. In contrast, the remaining layer blocks require much larger $\gamma$ to influence the output and still yield weaker gains. This pattern suggests that guidance is most effective in earlier layers, where cross-modal interactions are first established.

\subsection{Computational Efficiency}
\label{sec:efficiency}

As shown in Table~\ref{tab:efficiency}, we measure average wall-clock latency per image and per word on CHAIR (max new tokens $=128$, greedy decoding), using the same environment across methods. Our goal is to compare \emph{cost} (latency and number of forward passes) against \emph{benefit} (CHAIR$_i$).

Based on the observation in Sec.~\ref{sec:param_analysis} that early-layer guidance already yields substantial gains, we report two canonical operating modes:
\textbf{ACG-Full} (guidance on all layers) for maximum faithfulness, and \textbf{ACG-Fast} (guidance on the first 8 layers) as a more compute-conscious alternative.

Multi-pass baselines nearly double latency (1.97--2.28$\times$), whereas ACG remains single-pass. \textbf{ACG-Full} achieves the lowest CHAIR$_i$ (4.8) at only 1.19$\times$ the vanilla cost, outperforming the 2-pass PAI baseline (CHAIR$_i{=}7.6$) in both accuracy and speed. \textbf{ACG-Fast} retains most of the gains at near-vanilla cost (1.05$\times$). These results support using ACG-Full as the default setting and ACG-Fast as a compute-friendly alternative.

% \subsection{Qualitative Analysis}
% \label{sec:qualitative}
% In \cref{fig:qualitative sample}, we compare the responses on MMHal-Bench between the LLaVA-1.5 model and LLaVA-1.5 with ACG. In the upper example, LLaVA-1.5 hallucinates in the later decoding stages, whereas LLaVA-1.5 with ACG avoids such hallucination and instead correctly mentions the ``\texttt{brick floor}”. In the bottom example, LLaVA-1.5 shows an inaccurate scene interpretation and hallucination.
% % , stating ``\texttt{This photo is taken in a restaurant}” and hallucinating by saying ``\texttt{the presence of a dining table}”. 
% In contrast, LLaVA-1.5 with ACG correctly identifies that ``\texttt{This photo is taken at a beach}”, which matches with the ground truth answer.
% These qualitative examples validate that our method effectively suppresses hallucinations by preserving visual grounding and maintaining contextual accuracy throughout the generation process.
\subsection{Qualitative Analysis}
\label{sec:qualitative}

\begin{figure}[t]
  \centering
  \includegraphics[width=1.0\columnwidth]{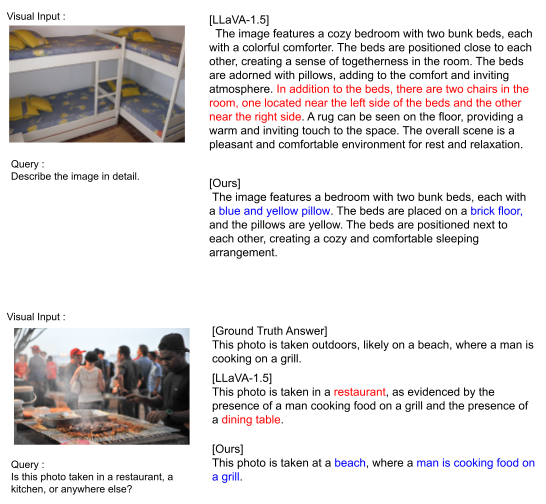}
\caption{\textbf{Comparison of responses from the baseline LLaVA-1.5 and LLaVA-1.5 with ACG.} Hallucinated or incorrect content is shown in \textcolor{red}{red}, and accurate content in \textcolor{blue}{blue}.}
  \label{fig:qualitative sample}
\end{figure}

In \cref{fig:qualitative sample}, we compare responses from the baseline LLaVA-1.5 model and LLaVA-1.5 with ACG on MMHal-Bench. In the upper example, the baseline hallucinates in the later decoding stages, whereas ACG avoids this error and instead correctly mentions the ``\texttt{brick floor}''. In the lower example, the baseline produces an inaccurate scene interpretation, while ACG correctly identifies that ``\texttt{This photo is taken at a beach}''.
These examples illustrate that ACG can suppress hallucinations while preserving visual grounding and improving contextual consistency.
\section{Conclusion}
\label{sec:conclusion}

In this paper, we proposed Attention-space Contrastive Guidance (ACG), a training-free, inference-time method for mitigating hallucinations in large vision--language models. ACG operates directly in self-attention layers by constructing an approximate text-only path via masked visual keys, and applying a textual orthogonalization step that suppresses components aligned with the text-only direction. Experiments on CHAIR and POPE show that ACG improves faithfulness over existing training-free baselines while remaining single-pass and computationally efficient.

More broadly, our results suggest that attention-space guidance is a practical and easily adoptable direction for improving image-grounded generation in LVLMs. Although ACG remains a simple inference-time intervention and does not fully resolve the broader problem of multimodal hallucination, it shows that directly reshaping attention toward visual evidence can substantially improve faithfulness at low additional cost. We hope this perspective encourages further work on lightweight, inference-time methods that make LVLMs attend more reliably to the image.
% In this paper, we proposed Attention-space Contrastive Guidance (ACG), a training-free, inference-time method for mitigating hallucinations in large vision--language models. ACG operates directly in self-attention layers by approximating a text-only path via masked visual keys and applying a textual orthogonalization step to isolate purely visual contributions geometrically. Experiments on CHAIR, POPE show that ACG attains state-of-the-art faithfulness while remaining single-pass and computationally efficient, outperforming prior multi-pass logit-level guidance methods in both accuracy and latency.

\newpage
\paragraph{Acknowledgements.}
This work was supported by Institute of Information \& communications Technology Planning \& Evaluation(IITP) under the Leading Generative AI Human Resources Development(IITP-2026-RS-2024-00397085) grant funded by the Korea government(MSIT).
This work was also supported by the National Research Foundation of Korea (NRF) grant funded by the Korea government (MSIT) (No. RS-2024-00345809, Research on AI Robustness Against Distribution Shift in Real-World Scenarios; and No. RS-2023-00222663, Center for Optimizing Hyperscale AI Models and Platforms).


{
    \small
    \bibliographystyle{ieeenat_fullname}
    \bibliography{main}

@String(CVPR= {IEEE Conf. Comput. Vis. Pattern Recog.})

@String(CVPR  = {CVPR})

@article{yue2023mmmu,
  title={MMMU: A Massive Multi-discipline Multimodal Understanding and Reasoning Benchmark for Expert AGI},
  year={2023},
  url={https://arxiv.org/abs/2311.16502}
}

@misc{lu2024mathvistaevaluatingmathematicalreasoning,
      title={MathVista: Evaluating Mathematical Reasoning of Foundation Models in Visual Contexts}, 
      year={2024},
      eprint={2310.02255},
      archivePrefix={arXiv},
      primaryClass={cs.CV},
      url={https://arxiv.org/abs/2310.02255}, 
}

@InProceedings{llava1_5,
    author    = {Liu, Haotian and Li, Chunyuan and Li, Yuheng and Lee, Yong Jae},
    title     = {Improved Baselines with Visual Instruction Tuning},
    booktitle = {Proceedings of the IEEE/CVF Conference on Computer Vision and Pattern Recognition (CVPR)},
    month     = {June},
    year      = {2024},
    pages     = {26296-26306}
}

@misc{zhu2023minigpt4enhancingvisionlanguageunderstanding,
      title={MiniGPT-4: Enhancing Vision-Language Understanding with Advanced Large Language Models}, 
      author={Deyao Zhu and Jun Chen and Xiaoqian Shen and Xiang Li and Mohamed Elhoseiny},
      year={2023},
      eprint={2304.10592},
      archivePrefix={arXiv},
      primaryClass={cs.CV},
      url={https://arxiv.org/abs/2304.10592}, 
}

@misc{bai2023qwenvlversatilevisionlanguagemodel,
      title={Qwen-VL: A Versatile Vision-Language Model for Understanding, Localization, Text Reading, and Beyond}, 
      author={Jinze Bai and Shuai Bai and Shusheng Yang and Shijie Wang and Sinan Tan and Peng Wang and Junyang Lin and Chang Zhou and Jingren Zhou},
      year={2023},
      eprint={2308.12966},
      archivePrefix={arXiv},
      primaryClass={cs.CV},
      url={https://arxiv.org/abs/2308.12966}, 
}

@misc{dai2023instructblipgeneralpurposevisionlanguagemodels,
      title={InstructBLIP: Towards General-purpose Vision-Language Models with Instruction Tuning}, 
      author={Wenliang Dai and Junnan Li and Dongxu Li and Anthony Meng Huat Tiong and Junqi Zhao and Weisheng Wang and Boyang Li and Pascale Fung and Steven Hoi},
      year={2023},
      eprint={2305.06500},
      archivePrefix={arXiv},
      primaryClass={cs.CV},
      url={https://arxiv.org/abs/2305.06500}, 
}

@misc{li2023blip2bootstrappinglanguageimagepretraining,
      title={BLIP-2: Bootstrapping Language-Image Pre-training with Frozen Image Encoders and Large Language Models}, 
      author={Junnan Li and Dongxu Li and Silvio Savarese and Steven Hoi},
      year={2023},
      eprint={2301.12597},
      archivePrefix={arXiv},
      primaryClass={cs.CV},
      url={https://arxiv.org/abs/2301.12597}, 
}

@misc{alayrac2022flamingovisuallanguagemodel,
      title={Flamingo: a Visual Language Model for Few-Shot Learning}, 
      author={Jean-Baptiste Alayrac and Jeff Donahue and Pauline Luc and Antoine Miech and Iain Barr and Yana Hasson and Karel Lenc and Arthur Mensch and Katie Millican and Malcolm Reynolds and Roman Ring and Eliza Rutherford and Serkan Cabi and Tengda Han and Zhitao Gong and Sina Samangooei and Marianne Monteiro and Jacob Menick and Sebastian Borgeaud and Andrew Brock and Aida Nematzadeh and Sahand Sharifzadeh and Mikolaj Binkowski and Ricardo Barreira and Oriol Vinyals and Andrew Zisserman and Karen Simonyan},
      year={2022},
      eprint={2204.14198},
      archivePrefix={arXiv},
      primaryClass={cs.CV},
      url={https://arxiv.org/abs/2204.14198}, 
}

@misc{jiang2024hallucinationaugmentedcontrastivelearning,
      title={Hallucination Augmented Contrastive Learning for Multimodal Large Language Model}, 
      author={Chaoya Jiang and Haiyang Xu and Mengfan Dong and Jiaxing Chen and Wei Ye and Ming Yan and Qinghao Ye and Ji Zhang and Fei Huang and Shikun Zhang},
      year={2024},
      eprint={2312.06968},
      archivePrefix={arXiv},
      primaryClass={cs.CV},
      url={https://arxiv.org/abs/2312.06968}, 
}

@misc{wu2025mitigatinghallucinationslargevisionlanguage,
      title={Mitigating Hallucinations in Large Vision-Language Models via Entity-Centric Multimodal Preference Optimization}, 
      author={Jiulong Wu and Zhengliang Shi and Shuaiqiang Wang and Jizhou Huang and Dawei Yin and Lingyong Yan and Min Cao and Min Zhang},
      year={2025},
      eprint={2506.04039},
      archivePrefix={arXiv},
      primaryClass={cs.CV},
      url={https://arxiv.org/abs/2506.04039}, 
}

@inproceedings{cd,
    title = "Contrastive Decoding: Open-ended Text Generation as Optimization",
    author = "Li, Xiang Lisa  and
      Holtzman, Ari  and
      Fried, Daniel  and
      Liang, Percy  and
      Eisner, Jason  and
      Hashimoto, Tatsunori  and
      Zettlemoyer, Luke  and
      Lewis, Mike",
    editor = "Rogers, Anna  and
      Boyd-Graber, Jordan  and
      Okazaki, Naoaki",
    booktitle = "Proceedings of the 61st Annual Meeting of the Association for Computational Linguistics (Volume 1: Long Papers)",
    month = jul,
    year = "2023",
    address = "Toronto, Canada",
    publisher = "Association for Computational Linguistics",
    url = "https://aclanthology.org/2023.acl-long.687/",
    doi = "10.18653/v1/2023.acl-long.687",
    pages = "12286--12312",
}

@InProceedings{vcd,
    author    = {Leng, Sicong and Zhang, Hang and Chen, Guanzheng and Li, Xin and Lu, Shijian and Miao, Chunyan and Bing, Lidong},
    title     = {Mitigating Object Hallucinations in Large Vision-Language Models through Visual Contrastive Decoding},
    booktitle = {Proceedings of the IEEE/CVF Conference on Computer Vision and Pattern Recognition (CVPR)},
    month     = {June},
    year      = {2024},
    pages     = {13872-13882}
}

@misc{lee2024delvevisualcontrastivedecoding,
      title={Delve into Visual Contrastive Decoding for Hallucination Mitigation of Large Vision-Language Models}, 
      author={Yi-Lun Lee and Yi-Hsuan Tsai and Wei-Chen Chiu},
      year={2024},
      eprint={2412.06775},
      archivePrefix={arXiv},
      primaryClass={cs.CV},
      url={https://arxiv.org/abs/2412.06775}, 
}

@misc{park2025secondmitigatingperceptualhallucination,
      title={SECOND: Mitigating Perceptual Hallucination in Vision-Language Models via Selective and Contrastive Decoding}, 
      author={Woohyeon Park and Woojin Kim and Jaeik Kim and Jaeyoung Do},
      year={2025},
      eprint={2506.08391},
      archivePrefix={arXiv},
      primaryClass={cs.CV},
      url={https://arxiv.org/abs/2506.08391}, 
}

@misc{icd,
      title={Mitigating Hallucinations in Large Vision-Language Models with Instruction Contrastive Decoding}, 
      author={Xintong Wang and Jingheng Pan and Liang Ding and Chris Biemann},
      year={2024},
      eprint={2403.18715},
      archivePrefix={arXiv},
      primaryClass={cs.CV},
      url={https://arxiv.org/abs/2403.18715}, 
}

@misc{chuang2024doladecodingcontrastinglayers,
      title={DoLa: Decoding by Contrasting Layers Improves Factuality in Large Language Models}, 
      author={Yung-Sung Chuang and Yujia Xie and Hongyin Luo and Yoon Kim and James Glass and Pengcheng He},
      year={2024},
      eprint={2309.03883},
      archivePrefix={arXiv},
      primaryClass={cs.CL},
      url={https://arxiv.org/abs/2309.03883}, 
}

@misc{wan2025onlyonelayerinterventionsufficiently,
      title={ONLY: One-Layer Intervention Sufficiently Mitigates Hallucinations in Large Vision-Language Models}, 
      author={Zifu Wan and Ce Zhang and Silong Yong and Martin Q. Ma and Simon Stepputtis and Louis-Philippe Morency and Deva Ramanan and Katia Sycara and Yaqi Xie},
      year={2025},
      eprint={2507.00898},
      archivePrefix={arXiv},
      primaryClass={cs.CV},
      url={https://arxiv.org/abs/2507.00898}, 
}

@misc{wan2024contrastiveregionguidanceimproving,
      title={Contrastive Region Guidance: Improving Grounding in Vision-Language Models without Training}, 
      author={David Wan and Jaemin Cho and Elias Stengel-Eskin and Mohit Bansal},
      year={2024},
      eprint={2403.02325},
      archivePrefix={arXiv},
      primaryClass={cs.CV},
      url={https://arxiv.org/abs/2403.02325}, 
}

@misc{ho2022classifierfreediffusionguidance,
      title={Classifier-Free Diffusion Guidance}, 
      author={Jonathan Ho and Tim Salimans},
      year={2022},
      eprint={2207.12598},
      archivePrefix={arXiv},
      primaryClass={cs.LG},
      url={https://arxiv.org/abs/2207.12598}, 
}

@misc{sanchez2023staytopicclassifierfreeguidance,
      title={Stay on topic with Classifier-Free Guidance}, 
      author={Guillaume Sanchez and Honglu Fan and Alexander Spangher and Elad Levi and Pawan Sasanka Ammanamanchi and Stella Biderman},
      year={2023},
      eprint={2306.17806},
      archivePrefix={arXiv},
      primaryClass={cs.CL},
      url={https://arxiv.org/abs/2306.17806}, 
}

@misc{marine,
      title={Mitigating Object Hallucination in Large Vision-Language Models via Image-Grounded Guidance}, 
      author={Linxi Zhao and Yihe Deng and Weitong Zhang and Quanquan Gu},
      year={2025},
      eprint={2402.08680},
      archivePrefix={arXiv},
      primaryClass={cs.LG},
      url={https://arxiv.org/abs/2402.08680}, 
}

@misc{zhang2024prompthighlighterinteractivecontrol,
      title={Prompt Highlighter: Interactive Control for Multi-Modal LLMs}, 
      author={Yuechen Zhang and Shengju Qian and Bohao Peng and Shu Liu and Jiaya Jia},
      year={2024},
      eprint={2312.04302},
      archivePrefix={arXiv},
      primaryClass={cs.CV},
      url={https://arxiv.org/abs/2312.04302}, 
}

@misc{pai,
      title={Paying More Attention to Image: A Training-Free Method for Alleviating Hallucination in LVLMs}, 
      author={Shi Liu and Kecheng Zheng and Wei Chen},
      year={2024},
      eprint={2407.21771},
      archivePrefix={arXiv},
      primaryClass={cs.CV},
      url={https://arxiv.org/abs/2407.21771}, 
}

@inproceedings{subramani-etal-2022-extracting,
    title = "Extracting Latent Steering Vectors from Pretrained Language Models",
    author = "Subramani, Nishant  and
      Suresh, Nivedita  and
      Peters, Matthew",
    editor = "Muresan, Smaranda  and
      Nakov, Preslav  and
      Villavicencio, Aline",
    booktitle = "Findings of the Association for Computational Linguistics: ACL 2022",
    month = may,
    year = "2022",
    address = "Dublin, Ireland",
    publisher = "Association for Computational Linguistics",
    url = "https://aclanthology.org/2022.findings-acl.48/",
    doi = "10.18653/v1/2022.findings-acl.48",
    pages = "566--581",
}

@inproceedings{su-etal-2025-activation,
    title = "Activation Steering Decoding: Mitigating Hallucination in Large Vision-Language Models through Bidirectional Hidden State Intervention",
    author = "Su, Jingran  and
      Chen, Jingfan  and
      Li, Hongxin  and
      Chen, Yuntao  and
      Qing, Li  and
      Zhang, Zhaoxiang",
    editor = "Che, Wanxiang  and
      Nabende, Joyce  and
      Shutova, Ekaterina  and
      Pilehvar, Mohammad Taher",
    booktitle = "Proceedings of the 63rd Annual Meeting of the Association for Computational Linguistics (Volume 1: Long Papers)",
    month = jul,
    year = "2025",
    address = "Vienna, Austria",
    publisher = "Association for Computational Linguistics",
    url = "https://aclanthology.org/2025.acl-long.634/",
    doi = "10.18653/v1/2025.acl-long.634",
    pages = "12964--12974",
    ISBN = "979-8-89176-251-0",
}

@inproceedings{wu-etal-2025-sharp,
    title = "{SHARP}: Steering Hallucination in {LVLM}s via Representation Engineering",
    author = "Wu, Junfei  and
      Ding, Yue  and
      Liu, Guofan  and
      Xia, Tianze  and
      Huang, Ziyue  and
      Sui, Dianbo  and
      Liu, Qiang  and
      Wu, Shu  and
      Wang, Liang  and
      Tan, Tieniu",
    editor = "Christodoulopoulos, Christos  and
      Chakraborty, Tanmoy  and
      Rose, Carolyn  and
      Peng, Violet",
    booktitle = "Proceedings of the 2025 Conference on Empirical Methods in Natural Language Processing",
    month = nov,
    year = "2025",
    address = "Suzhou, China",
    publisher = "Association for Computational Linguistics",
    url = "https://aclanthology.org/2025.emnlp-main.725/",
    doi = "10.18653/v1/2025.emnlp-main.725",
    pages = "14357--14372",
    ISBN = "979-8-89176-332-6",
}

@misc{vti,
      title={Reducing Hallucinations in Vision-Language Models via Latent Space Steering}, 
      author={Sheng Liu and Haotian Ye and Lei Xing and James Zou},
      year={2024},
      eprint={2410.15778},
      archivePrefix={arXiv},
      primaryClass={cs.CV},
      url={https://arxiv.org/abs/2410.15778}, 
}

@InProceedings{vista,
  title = 	 {The Hidden Life of Tokens: Reducing Hallucination of Large Vision-Language Models Via Visual Information Steering},
  author =       {Li, Zhuowei and Shi, Haizhou and Gao, Yunhe and Liu, Di and Wang, Zhenting and Chen, Yuxiao and Liu, Ting and Zhao, Long and Wang, Hao and Metaxas, Dimitris N.},
  booktitle = 	 {Proceedings of the 42nd International Conference on Machine Learning},
  pages = 	 {35799--35819},
  year = 	 {2025},
  editor = 	 {Singh, Aarti and Fazel, Maryam and Hsu, Daniel and Lacoste-Julien, Simon and Berkenkamp, Felix and Maharaj, Tegan and Wagstaff, Kiri and Zhu, Jerry},
  volume = 	 {267},
  series = 	 {Proceedings of Machine Learning Research},
  month = 	 {13--19 Jul},
  publisher =    {PMLR},
  url = 	 {https://proceedings.mlr.press/v267/li25ca.html},
}

@inproceedings{
adhh,
title={Mitigating Hallucination in Large Vision-Language Models via Modular Attribution and Intervention},
author={Tianyun Yang and Ziniu Li and Juan Cao and Chang Xu},
booktitle={The Thirteenth International Conference on Learning Representations},
year={2025},
url={https://openreview.net/forum?id=Bjq4W7P2Us}
}

@misc{jung2025visualattentionfadesselective,
      title={Visual Attention Never Fades: Selective Progressive Attention ReCalibration for Detailed Image Captioning in Multimodal Large Language Models}, 
      author={Mingi Jung and Saehyung Lee and Eunji Kim and Sungroh Yoon},
      year={2025},
      eprint={2502.01419},
      archivePrefix={arXiv},
      primaryClass={cs.CV},
      url={https://arxiv.org/abs/2502.01419}, 
}

@misc{kang2025toldvisualattentionsink,
      title={See What You Are Told: Visual Attention Sink in Large Multimodal Models}, 
      author={Seil Kang and Jinyeong Kim and Junhyeok Kim and Seong Jae Hwang},
      year={2025},
      eprint={2503.03321},
      archivePrefix={arXiv},
      primaryClass={cs.CV},
      url={https://arxiv.org/abs/2503.03321}, 
}

@misc{jiang2025devilsmiddlelayerslarge,
      title={Devils in Middle Layers of Large Vision-Language Models: Interpreting, Detecting and Mitigating Object Hallucinations via Attention Lens}, 
      author={Zhangqi Jiang and Junkai Chen and Beier Zhu and Tingjin Luo and Yankun Shen and Xu Yang},
      year={2025},
      eprint={2411.16724},
      archivePrefix={arXiv},
      primaryClass={cs.CV},
      url={https://arxiv.org/abs/2411.16724}, 
}

@misc{chen2025spatialreasoninghardvlms,
      title={Why Is Spatial Reasoning Hard for VLMs? An Attention Mechanism Perspective on Focus Areas}, 
      author={Shiqi Chen and Tongyao Zhu and Ruochen Zhou and Jinghan Zhang and Siyang Gao and Juan Carlos Niebles and Mor Geva and Junxian He and Jiajun Wu and Manling Li},
      year={2025},
      eprint={2503.01773},
      archivePrefix={arXiv},
      primaryClass={cs.CL},
      url={https://arxiv.org/abs/2503.01773}, 
}

@misc{zhu2025mitigatingobjecthallucinationslarge,
      title={Mitigating Object Hallucinations in Large Vision-Language Models via Attention Calibration}, 
      author={Younan Zhu and Linwei Tao and Minjing Dong and Chang Xu},
      year={2025},
      eprint={2502.01969},
      archivePrefix={arXiv},
      primaryClass={cs.CV},
      url={https://arxiv.org/abs/2502.01969}, 
}

@inproceedings{pope,
    title = "Evaluating Object Hallucination in Large Vision-Language Models",
    author = "Li, Yifan  and
      Du, Yifan  and
      Zhou, Kun  and
      Wang, Jinpeng  and
      Zhao, Xin  and
      Wen, Ji-Rong",
    editor = "Bouamor, Houda  and
      Pino, Juan  and
      Bali, Kalika",
    booktitle = "Proceedings of the 2023 Conference on Empirical Methods in Natural Language Processing",
    month = dec,
    year = "2023",
    address = "Singapore",
    publisher = "Association for Computational Linguistics",
    url = "https://aclanthology.org/2023.emnlp-main.20/",
    doi = "10.18653/v1/2023.emnlp-main.20",
    pages = "292--305",
}

@inproceedings{chair,
    title = "Object Hallucination in Image Captioning",
    author = "Rohrbach, Anna  and
      Hendricks, Lisa Anne  and
      Burns, Kaylee  and
      Darrell, Trevor  and
      Saenko, Kate",
    editor = "Riloff, Ellen  and
      Chiang, David  and
      Hockenmaier, Julia  and
      Tsujii, Jun{'}ichi",
    booktitle = "Proceedings of the 2018 Conference on Empirical Methods in Natural Language Processing",
    month = oct # "-" # nov,
    year = "2018",
    address = "Brussels, Belgium",
    publisher = "Association for Computational Linguistics",
    url = "https://aclanthology.org/D18-1437/",
    doi = "10.18653/v1/D18-1437",
    pages = "4035--4045",
}

@inproceedings{mmhal,
    title = "Aligning Large Multimodal Models with Factually Augmented {RLHF}",
    author = "Sun, Zhiqing  and
      Shen, Sheng  and
      Cao, Shengcao  and
      Liu, Haotian  and
      Li, Chunyuan  and
      Shen, Yikang  and
      Gan, Chuang  and
      Gui, Liangyan  and
      Wang, Yu-Xiong  and
      Yang, Yiming  and
      Keutzer, Kurt  and
      Darrell, Trevor",
    editor = "Ku, Lun-Wei  and
      Martins, Andre  and
      Srikumar, Vivek",
    booktitle = "Findings of the Association for Computational Linguistics: ACL 2024",
    month = aug,
    year = "2024",
    address = "Bangkok, Thailand",
    publisher = "Association for Computational Linguistics",
    url = "https://aclanthology.org/2024.findings-acl.775/",
    doi = "10.18653/v1/2024.findings-acl.775",
    pages = "13088--13110",
}

@misc{kaul2025throneobjectbasedhallucinationbenchmark,
      title={THRONE: An Object-based Hallucination Benchmark for the Free-form Generations of Large Vision-Language Models}, 
      author={Prannay Kaul and Zhizhong Li and Hao Yang and Yonatan Dukler and Ashwin Swaminathan and C. J. Taylor and Stefano Soatto},
      year={2025},
      eprint={2405.05256},
      archivePrefix={arXiv},
      primaryClass={cs.CV},
      url={https://arxiv.org/abs/2405.05256}, 
}

@misc{bai2025hallucinationmultimodallargelanguage,
      title={Hallucination of Multimodal Large Language Models: A Survey}, 
      author={Zechen Bai and Pichao Wang and Tianjun Xiao and Tong He and Zongbo Han and Zheng Zhang and Mike Zheng Shou},
      year={2025},
      eprint={2404.18930},
      archivePrefix={arXiv},
      primaryClass={cs.CV},
      url={https://arxiv.org/abs/2404.18930}, 
}

@misc{zuo2025medhallbenchnewbenchmarkassessing,
      title={MedHallBench: A New Benchmark for Assessing Hallucination in Medical Large Language Models}, 
      author={Kaiwen Zuo and Yirui Jiang},
      year={2025},
      eprint={2412.18947},
      archivePrefix={arXiv},
      primaryClass={cs.CL},
      url={https://arxiv.org/abs/2412.18947}, 
}

@misc{petryk2024alohanewmeasurehallucination,
      title={ALOHa: A New Measure for Hallucination in Captioning Models}, 
      author={Suzanne Petryk and David M. Chan and Anish Kachinthaya and Haodi Zou and John Canny and Joseph E. Gonzalez and Trevor Darrell},
      year={2024},
      eprint={2404.02904},
      archivePrefix={arXiv},
      primaryClass={cs.CV},
      url={https://arxiv.org/abs/2404.02904}, 
}
}
\clearpage
\setcounter{page}{1}
\maketitlesupplementary
\appendix

\section{Additional Experimental Details}
\label{sec:impl_details}
\subsection{Models.}
We evaluate ACG on three open-source large vision--language models (LVLMs) with diverse
language backbones and vision--language connectors:
LLaVA-1.5, MiniGPT-4, and Qwen-VL-Chat.

\paragraph{LLaVA-1.5.~\cite{llava1_5}}
LLaVA-1.5 adopts a CLIP ViT-L/336px vision encoder and a Vicuna language model built on the LLaMA architecture, connected by a fully connected MLP-based vision--language projector.
The CLIP encoder produces 576 visual tokens per image, which are mapped into the language model token embedding space by a two-layer MLP and then concatenated with text tokens.
The model is trained in a two-stage pipeline: vision--language alignment on image--text pairs, followed by visual instruction tuning on conversational multimodal data.

\paragraph{MiniGPT-4.~\cite{zhu2023minigpt4enhancingvisionlanguageunderstanding}}
MiniGPT-4 uses the visual frontend of BLIP-2: a ViT-G/14 visual encoder from EVA-CLIP followed by a Q-Former that compresses dense image features into a small set of visual tokens.
The Q-Former employs a fixed set of learnable queries (32 in our setup), so each image is represented as 32 visual tokens.
These Q-Former outputs are passed through a single linear projection layer to align them with the Vicuna language model embedding space, and the projected visual tokens are then fed into Vicuna as a soft prompt for generation.

\paragraph{Qwen-VL-Chat.~\cite{bai2023qwenvlversatilevisionlanguagemodel}}
Qwen-VL-Chat builds on the Qwen language model and a ViT-bigG visual encoder from OpenCLIP.
Image features are first extracted by the ViT encoder and then compressed by a position-aware vision--language adapter: a single-layer cross-attention module with learnable query embeddings.
We use the default configuration with 256 queries, yielding 256 visual tokens per image, which are fed into the language model as a fixed-length visual token block.

\subsection{Benchmarks.}
% POPE/CHAIR 좀 더 자세하게 (metric 수식 포함) -> metric, dataset split, 뭘 측정했고 뭐가 중요한지 자세하게, + MMHal  -> 상윤
\paragraph{POPE.~\cite{pope}}
POPE (Precision-based Object Probing Evaluation) evaluates object hallucination through binary object-presence queries. 
For each image, the model answers questions of the form ``\texttt{Is there a <object> in the image?}'' with a balanced mixture of present and absent objects. 
POPE provides three complementary evaluation sets:
(1) \textit{Random}: object categories are sampled uniformly from the vocabulary, reflecting unbiased hallucination performance; 
(2) \textit{Popular}: focuses on frequently occurring objects in large-scale training corpora, testing whether the model over-relies on language priors; 
(3) \textit{Adversarial}: selects semantically or visually confusable objects (e.g., querying ``cat'' for dog images), stressing context-induced hallucination.

\noindent\textbf{Metrics.}
POPE reports Accuracy, Precision, Recall, and F1.

\paragraph{CHAIR.~\cite{chair}}
CHAIR (Caption Hallucination Assessment with Image Re-annotation) measures hallucination in image captioning by aligning object mentions in generated captions with COCO ground-truth annotations. 
Any object mentioned in the caption but absent in the image is regarded as hallucinated.

\noindent\textbf{Metrics.}
CHAIR provides two metrics:
\[
\begin{aligned}
\text{CHAIR}_i = \frac{\#\text{hallucinated object instances}}{\#\text{all mentioned objects}}, \qquad \\
\text{CHAIR}_s = \frac{\#\text{captions containing hallucination}}{\#\text{total captions}}.
\end{aligned}
\]
CHAIR$_i$ captures object-level hallucination frequency, while CHAIR$_s$ measures how often a caption contains any hallucination. 

\noindent\textbf{Why F1 is Reported Alongside CHAIR$_s$ and CHAIR$_i$.}
CHAIR$_i$ and CHAIR$_s$ measure hallucination from the perspective of
``how often'' hallucinated objects appear in a caption, but they do not consider 
whether the model successfully mentions objects that actually exist in the image. 
In contrast, the object-level F1 score captures the balance between avoiding hallucinated objects (precision) and correctly mentioning ground-truth objects (recall). 
A model may achieve a low CHAIR score simply by producing overly conservative captions that omit many valid objects, which results in low recall.
Therefore, reporting F1 alongside CHAIR$_i$ and CHAIR$_s$ provides a more complete view of caption quality, distinguishing models that truly reduce hallucination from those that merely under-describe the image.

% \vspace{0.5em}
\paragraph{MMHal-Bench.~\cite{mmhal}}
MMHal-Bench is a hallucination-centric benchmark specifically designed to diagnose the visual grounding reliability of large vision–language models (LVLMs). 
The dataset contains images paired with carefully constructed natural-language queries that target reasoning types known to induce hallucination, such as object attributes, spatial relations, counting, and adversarially misleading premises. 
Unlike captioning-based or binary object-probing benchmarks, MMHal-Bench evaluates open-ended, reasoning-intensive responses, where hallucinations arise not only in object mentions but also in relational, numerical, or contextual inferences. 
Each query expects a concise, grounded answer that can be automatically judged for hallucination and informativeness using an LLM-as-a-judge protocol.

The benchmark evaluates hallucination across several reasoning dimensions that are known to induce failure in LVLMs:
\begin{itemize}
    \item \textbf{ATTR (Object Attributes):} questions about appearance attributes such as color, shape, texture, or material (e.g., ``What color is the man's jacket?''). 
    \item \textbf{ADV (Adversarial Objects):} queries intentionally designed to include objects not present in the image (e.g., ``What is the dog holding in its hand?''). This category tests the model's robustness to adversarial wording and its ability to reject false premises.
    \item \textbf{COMP (Comparisons):} relational comparisons of size, number, or attributes between two or more objects (e.g., ``Which cup is larger?''). 
    \item \textbf{COUNT (Counting):} numerical reasoning about the number of instances. 
    \item \textbf{SPAT (Spatial Relations):} reasoning about object positions or geometric relations (e.g., ``Where is the bicycle relative to the car?''). 
    \item \textbf{ENV (Environmental / Scene Inference):} global contextual reasoning about the environment, scene type, or high-level situational cues (e.g., ``Is this an indoor or outdoor scene?'').
\end{itemize}

\noindent\textbf{Evaluation Protocol.}
MMHal adopts an LLM-as-a-judge evaluation pipeline in which GPT-4 scores each model response along two dimensions: \emph{informativeness} and \emph{hallucination}. 
For every image--query pair, GPT-4 is prompted to evaluate (1) how informative the response is on a 0--6 scale, where 0 indicates an unhelpful or irrelevant answer and 6 indicates a fully grounded, complete, and contextually appropriate answer; and (2) whether the response contains any hallucinated visual content (binary judgment: hallucinated / not hallucinated). 
This protocol enables the benchmark to assess both the usefulness and the visual faithfulness of a model's answer.

\subsection{Baseline Methods and Hyperparameters}
% - VCD / PAI / VISTA 구현 및 튜닝 설정 / 공식 repo 코드 기반, 저자들이 지정한 하이퍼파라미터로 돌렸음(VISTA 제외) 등등
 To demonstrate our method's generability among tasks, we set $\gamma $ to 2.4 for LLaVA-1.5~\cite{llava1_5}, 0.3 for MiniGPT-4~\cite{zhu2023minigpt4enhancingvisionlanguageunderstanding}, 1.4 for Qwen-VL~\cite{bai2023qwenvlversatilevisionlanguagemodel} for all experiments unless explicitly stated otherwise. For a consistent and comparative analysis, we set greedy decoding as default. 

For baseline comparisons, we evaluated VCD~\cite{vcd}, PAI~\cite{pai}, and VISTA~\cite{vista}. All baselines were reproduced using their official code repositories, and all experiments were conducted under a unified greedy decoding setting. For hyperparameters, we followed the configurations reported in the original papers unless otherwise noted. An exception is VISTA: as prior reports indicate that the official hyperparameters do not reproduce the reported results, we conducted a parameter search and set \texttt{vsv-lambda} to 0.01 for POPE and 0.15 for CHAIR.

\section{Quantitative Validation of ACG}
\subsection{Justifying the Masked Unconditional Path.}
We quantitatively validate whether our \textbf{single-pass} masked text-only path matches a true image-absent forward pass.
On 500 samples from a random POPE split with LLaVA-1.5, 
We form \(\tilde{U}\) by running one image-conditioned forward pass, masking \emph{image keys} for the \emph{last generated token} at each layer, and propagating the masked outputs across layers.
We compute the true text-only trajectory \(U\) via a separate forward pass without images.
Using attention distributions and logits, \(\tilde{U}\) closely tracks \(U\) throughout the network (Table~\ref{tab:approx_quality_sub}), suggesting that any residual leakage/redistribution remains bounded and does not induce a divergent text-only path.

\begin{table}[t]
\centering
\caption{\textbf{ACG quantitative validation.} (LLaVA-1.5, POPE random split)}
\label{tab:approx_and_orth}
\scriptsize
\begin{subtable}[t]{\columnwidth}
\centering
\caption{\textbf{Text-only path approximation quality.} We compare $\tilde U$ to the true text-only trajectory $U$ using attention distributions (Attn) and logits.}
\label{tab:approx_quality_sub}
\setlength{\tabcolsep}{3.5pt} % tighten to avoid overflow
\begin{tabular}{lcccc}
\toprule
 & Attn cos \(\uparrow\) & Attn KL \(\downarrow\) & Logits cos \(\uparrow\) & Top-10 overlap \(\uparrow\) \\
\midrule
\(\tilde{U}\) vs.\ \(U\) & 0.935 & 0.195 & 0.909 & 0.705 \\
\bottomrule
\end{tabular}
\end{subtable}

\begin{subtable}[t]{\columnwidth}
\centering
\caption{\textbf{Orthogonalization effect.} $z_C$ conditional logits, $z_U$ true text-only logits, and $z_{\tilde U}$ logits from the single-pass masked-and-propagated trajectory.}
\label{tab:orth_effect_core_sub}
\scriptsize
\setlength{\tabcolsep}{3.6pt}
\renewcommand{\arraystretch}{1.05}
\begin{tabular}{lccc}
\toprule
 & \(\Delta_{\text{mask}}(=z_C-z_{\tilde U})\) & ACG no-orth & ACG + orth \\
\midrule
\multicolumn{4}{l}{\textbf{Mechanism (attention-output level, \(\Delta O\))}} \\
proj\_ratio\((\Delta O\!\rightarrow\!\text{text})\)\(\downarrow\) & -- & 0.407 & \(2\times10^{-4}\) \\
\midrule
\multicolumn{4}{l}{\textbf{Behavior (logits level, \(\Delta z\))}} \\
cos\((\Delta z,\Delta_{\text{true}})\)\(\uparrow\) & 0.649 & 0.469 & 0.592 \\
proj\_ratio\((\Delta z\!\rightarrow\! z_U)\)\(\downarrow\) & 0.591 & 0.790 & 0.577 \\
\bottomrule
\end{tabular}

\end{subtable}
\end{table}

\subsection{Effect of textual orthogonalization.}
Textual Orthogonalization removes the text-direction component of the layerwise steering signal, and we verify this mechanism yields improved steering behavior at the logits level (Table~\ref{tab:orth_effect_core_sub}).
At the attention-output level, orthogonalization effectively eliminates the text-direction projection.
At the logits level, it increases alignment to the true steering direction and reduces text-only leakage measured by proj\_ratio\((\Delta z \rightarrow z_U)\).
Overall, orthogonalization acts as a targeted correction of text-bias in the steering signal that empirically restores the desired steering behavior.

% - γ sweep 테이블 및 canonical γ 선택 기준
\begin{table}[t]
\centering
\caption{\textbf{Effect of guidance scale $\gamma$ on CHAIR (max 128) for LLaVA-1.5.}
We report sentence-level hallucination (CHAIR$_s$), instance-level hallucination (CHAIR$_i$), F1, and average caption length (Len).
We choose $\gamma=\mathbf{2.4}$ as our operating point.}
\label{tab:gamma_llava}
\setlength{\tabcolsep}{4pt}
\begin{tabular}{ccccc}
\toprule
$\gamma$ & CHAIR$_s$ (↓) & CHAIR$_i$ (↓) & F1 (↑) & Len \\
\midrule
1.0 & 47.4 & 12.8 & 77.8 & 91.6 \\
% 1.2 & 49.6 & 12.9 & 77.7 & 90.6 \\
1.3 & 48.6 & 12.9 & 78.1 & 89.4 \\
1.5 & 44.8 & 11.6 & 78.1 & 88.2 \\
% 1.6 & 42.6 & 10.4 & 78.7 & 86.4 \\
1.7 & 42.4 & 10.3 & 79.1 & 84.9 \\
% 1.8 & 38.8 &  9.6 & 79.0 & 85.1 \\
1.9 & 36.4 &  8.6 & 79.3 & 83.5 \\
2.0 & 33.0 &  8.0 & 79.0 & 82.1 \\
2.1 & 34.2 &  7.6 & 77.6 & 80.8 \\
2.2 & 27.6 &  6.4 & 76.7 & 77.5 \\
2.3 & 27.8 &  6.6 & 75.5 & 74.9 \\
\textbf{2.4} & \textbf{21.0} & \textbf{4.8} & \textbf{74.4} & 72.4 \\
2.5 & 19.2 &  4.8 & 72.4 & 67.2 \\
2.6 & 15.4 &  4.8 & 68.0 & 58.7 \\
2.7 & 12.8 &  4.9 & 64.7 & 51.0 \\
2.8 &  9.0 &  4.8 & 60.4 & 42.0 \\
2.9 &  7.2 &  6.0 & 56.1 & 33.9 \\
3.0 &  6.2 &  4.2 & 51.8 & 25.8 \\
\bottomrule
\end{tabular}
\end{table}

\begin{table}[t]
\centering
\caption{\textbf{Effect of guidance scale $\gamma$ on CHAIR (max 128) for MiniGPT-4.}
We again report CHAIR$_s$, CHAIR$_i$, F1, and average caption length (Len).
We choose $\gamma=\mathbf{0.3}$ as our operating point.}
\label{tab:gamma_minigpt4}
\setlength{\tabcolsep}{4pt}
\begin{tabular}{ccccc}
\toprule
$\gamma$ & CHAIR$_s$ (↓) & CHAIR$_i$ (↓) & F1 (↑) & Len \\
\midrule
% 0.05 & 30.6 &  9.6 & 70.5 & 78.4 \\
0.10 & 27.6 &  9.0 & 70.6 & 72.9 \\
0.15 & 24.4 &  7.9 & 71.0 & 69.4 \\
0.20 & 21.0 &  6.8 & 70.3 & 63.5 \\
0.25 & 16.6 &  5.2 & 69.5 & 59.1 \\
\textbf{0.30} & \textbf{10.8} & \textbf{3.3} & \textbf{68.0} & 66.3 \\
0.35 &  6.2 &  2.5 & 63.2 & 73.6 \\
0.40 & 2.6 &  1.6 & 54.3 & 94.9 \\

\bottomrule
\end{tabular}
\end{table}

\begin{table*}[t]
\centering
\caption{\textbf{Layer-block ablation on LLaVA-1.5 (CHAIR, max 128).}
We apply ACG only to a given layer block and report CHAIR$_s$, CHAIR$_i$, F1, and average caption length (Len).}
\label{tab:layerblock_llava}
\setlength{\tabcolsep}{4pt}

\begin{minipage}{0.48\linewidth}
\centering
\caption*{(a) Early (layers 1--8, ACG-Fast)}
\begin{tabular}{ccccc}
\toprule
$\gamma$ & CHAIR$_s$ & CHAIR$_i$ & F1 & Len \\
\midrule
0.5 & 47.8 & 12.6 & 77.1 & 93.4 \\
1.0 & 46.0 & 13.1 & 77.0 & 91.9 \\
1.5 & 43.2 & 12.2 & 77.6 & 88.3 \\
2.0 & 35.8 & 10.2 & 77.4 & 82.2 \\
2.5 & 28.0 &  7.1 & 77.5 & 77.6 \\
3.0 & 19.0 &  5.3 & 69.8 & 68.0 \\
6.0 &  0.2 &  2.9 & 13.2 & 15.6 \\
10.0&  0.0 &  0.0 &  0.2 &  1.2 \\
\bottomrule
\end{tabular}
\end{minipage}
\hfill
\begin{minipage}{0.48\linewidth}
\centering
\caption*{(b) Early–mid (layers 9--16)}
\begin{tabular}{ccccc}
\toprule
$\gamma$ & CHAIR$_s$ & CHAIR$_i$ & F1 & Len \\
\midrule
0.5 & 49.2 & 14.0 & 76.6 & 94.3 \\
1.0 & 49.0 & 14.2 & 76.4 & 93.7 \\
2.0 & 53.6 & 14.4 & 75.9 & 92.4 \\
2.5 & 51.0 & 13.9 & 76.3 & 92.0 \\
3.0 & 53.6 & 15.3 & 74.8 & 90.7 \\
6.0 & 38.0 & 11.1 & 74.8 & 79.2 \\
8.0 & 14.2 &  5.7 & 56.6 & 45.9 \\
10.0&  2.2 &  2.1 & 22.5 & 16.3 \\
\bottomrule
\end{tabular}
\end{minipage}

\vspace{0.5em}

\begin{minipage}{0.48\linewidth}
\centering
\caption*{(c) Mid–late (layers 17--24)}
\begin{tabular}{ccccc}
\toprule
$\gamma$ & CHAIR$_s$ & CHAIR$_i$ & F1 & Len \\
\midrule
0.5 & 46.4 & 12.6 & 77.6 & 94.4 \\
1.0 & 46.4 & 12.5 & 78.3 & 93.4 \\
1.5 & 43.8 & 12.5 & 78.2 & 94.6 \\
2.5 & 47.2 & 12.7 & 78.1 & 95.6 \\
4.0 & 51.8 & 12.7 & 76.6 & 95.9 \\
6.0 & 42.6 & 10.7 & 77.4 & 95.6 \\
8.0 & 40.2 &  9.2 & 76.3 & 95.0 \\
10.0& 35.8 &  7.0 & 73.4 & 91.8 \\
\bottomrule
\end{tabular}
\end{minipage}
\hfill
\begin{minipage}{0.48\linewidth}
\centering
\caption*{(d) Late (layers 25--32)}
\begin{tabular}{ccccc}
\toprule
$\gamma$ & CHAIR$_s$ & CHAIR$_i$ & F1 & Len \\
\midrule
0.5 & 44.0 & 12.8 & 76.9 & 93.5 \\
1.0 & 43.8 & 12.2 & 77.4 & 93.7 \\
1.5 & 44.2 & 12.1 & 77.7 & 93.1 \\
2.0 & 41.8 & 11.4 & 78.4 & 93.6 \\
2.5 & 40.4 & 10.7 & 78.6 & 93.0 \\
4.0 & 39.4 & 10.2 & 78.9 & 92.5 \\
6.0 & 37.8 &  9.7 & 78.2 & 93.4 \\
10.0& 36.0 &  8.8 & 75.7 & 90.7 \\
\bottomrule
\end{tabular}
\end{minipage}
\end{table*}

\section{Behavior Across Guidance Scale and Depth}
\label{sec:ablation}
\subsection{Guidance Scale Selection.}
\label{sec:gamma_selection}
For each LVLM, we select the guidance scale $\gamma$ on the CHAIR (max 128 tokens) benchmark by sweeping $\gamma$ and monitoring the trade-off between hallucination and caption quality.
Concretely, we measure sentence-level hallucination (CHAIR$_s$), instance-level hallucination (CHAIR$_i$), F1, and the average caption length (Len) under greedy decoding, and choose an operating point that (i) substantially reduces CHAIR$_i$ compared to the greedy baseline, while (ii) keeping F1 within roughly $5\%$ of the baseline and (iii) avoiding degenerate overly short captions.
The selected $\gamma$ is then reused for all CHAIR(both max tokens 64 and 128) and POPE experiments on the corresponding model.

This protocol is consistent with prior training-free hallucination mitigation methods. For instance, PAI~\cite{pai} tunes its scaling parameters by sweeping on the CHAIR benchmark itself, jointly considering CHAIR and F1, without introducing a separate validation split.

Table~\ref{tab:gamma_llava} shows the sweep for LLaVA-1.5.
As $\gamma$ increases from $1.0$ to $2.4$, CHAIR$_i$ consistently decreases from $12.8$ to $4.8$, while F1 only drops from $77.8$ to $74.4$ and the average length remains moderate (72.4 tokens).
Beyond $\gamma=2.4$, CHAIR$_i$ continues to decrease but F1 and length collapse sharply (e.g., F1 $=51.8$ and Len $=25.8$ at $\gamma=3.0$), indicating an over-aggressive regime.
We therefore choose $\gamma=2.4$ as the canonical operating point for LLaVA-1.5. A similar trend is observed for MiniGPT-4 in Table~\ref{tab:gamma_minigpt4}.
We select $\gamma=0.3$, while avoiding the more unstable behavior at larger $\gamma$.

\begin{table*}[t]
\centering
\caption{\textbf{Category-wise informativeness and hallucination rate on MMHal.}}
\label{tab:mmhal}
\small
\begin{tabular}{lcccccccccc}
\toprule
$\gamma$ & Avg & ATTR & ADV & COMP & COUNT & SPAT & ENV & HOL & OTH & Hallucination Rate \\
\midrule
0 (Vanilla) & 1.94 & 2.25 & 1.42 & 2.67 & 1.50 & 1.92 & 2.92 & 1.75 & 1.08 & 0.59 \\
1 & 2.06 & 2.25 & 1.33 & 2.17 & 1.33 & 2.75 & 3.25 & 2.08 & 1.33 & 0.56 \\
1.5 & 2.07 & 2.50 & 1.08 & 1.75 & 1.92 & 2.00 & 3.25 & 2.08 & 2.00 & 0.57 \\
2 & 2.12 & 2.50 & 1.67 & 1.92 & 1.58 & 2.25 & 3.25 & 2.00 & 1.83 & 0.53 \\
2.4 & 2.01 & 2.58 & 2.00 & 1.83 & 1.17 & 2.25 & 3.25 & 1.17 & 1.83 & 0.56 \\
\bottomrule
\end{tabular}
\end{table*}

\subsection{Layer-block configurations.}
\label{sec:layerblock}
We also study where to apply ACG inside the LLaVA-1.5 decoder by restricting
the guidance to different layer blocks.
Table~\ref{tab:layerblock_llava} reports CHAIR$_s$, CHAIR$_i$, F1, and average
caption length (Len) on CHAIR (max 128) when we apply ACG only to
early (layers 1--8), early--mid (9--16), mid--late (17--24), or late (25--32) blocks.
We denote the early-only variant (layers 1--8) as \emph{ACG-Fast}, which
offers a good trade-off between hallucination reduction and efficiency.

\noindent\textbf{Interpretation.}
Early-layer guidance is notably more efficient than mid-to-late guidance:
it achieves strong hallucination reduction with smaller $\gamma$, whereas deeper blocks require substantially larger coefficients for comparable effects.
This trend is consistent with the view that early layers are a favorable intervention point for lightweight guidance, motivating our ACG-Fast variant.

\section{Limitations and Future Work}
\label{sec:limit_future}

We highlight two limitations of our method and several directions for future work.

\noindent\textbf{Architecture-specific assumptions.}
ACG assumes a standard LVLM design in which visual tokens appear as a contiguous block in the decoder input. However, architectures such as InstructBLIP employ Q-Former-based encoders or cross-attention modules that interleave visual information more tightly with the text stream. In such settings, simply masking visual-key positions may not fully remove visual information, because contextualized text embeddings can already contain fused vision features. Future work could develop architecture-aware masking strategies or alternative constructions of the text-only path that better match each model's multimodal fusion mechanism.

\noindent\textbf{Depth-dependent sensitivity.}
ACG also exhibits noticeable depth-dependent behavior: early layers respond strongly to guidance, whereas mid-to-late layers require significantly larger $\gamma$ to achieve comparable hallucination reduction. This suggests that a uniform guidance scale may under-correct or over-correct depending on the layer. More adaptive designs---such as depth-specific scaling, head-wise weighting, or learned guidance schedules---may further stabilize ACG across layers and improve hallucination mitigation.

\section{Additional Experimental Results and Qualitative Examples}
\label{sec:extended_results}
\subsection{Detailed MMHal-Bench Scores.}
In addition to the MMHal-Bench scores reported in main paper, we also provide results across multiple values of $\gamma$, ranging from 1.0 to 2.4. As shown in Table~\ref{tab:mmhal}, our method consistently achieves higher scores and lower hallucination rates than the vanilla model, achieving highest score at $\gamma=2$.

%%%%%%%%%%Qualatative Examples-MMHHal-Bench
\begin{figure*}[t]
    \centering
    \includegraphics[width=0.9\linewidth]{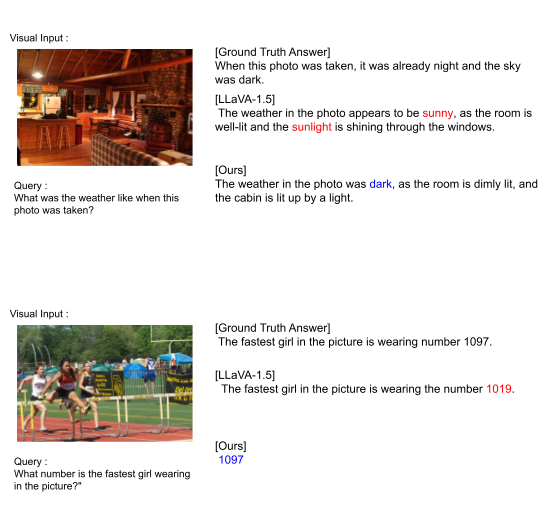}
    \caption{
    \textbf{Success examples on MMHal-Bench.}
    In the first example, vanilla LLaVA-1.5 hallucinates a bright and \textcolor{red}{sunny} environment, whereas our method correctly infers that the cabin is \textcolor{blue}{dark} and lit by artificial light.
    In the second example, vanilla misreads the runner's bib number as \textcolor{red}{1019}, while our method outputs the correct number \textcolor{blue}{1097}, matching the ground-truth answer.
    }
    \label{fig:mmhal_qual}
\end{figure*}

\subsection{Additional examples on MMHal-Bench.}
Figure~\ref{fig:mmhal_qual} provides additional examples from MMHal-Bench.
For the environmental reasoning query, vanilla LLaVA-1.5 describes the scene as \textcolor{red}{sunny} and well-lit, following language priors, whereas our method correctly answers that the weather appears \textcolor{blue}{dark} because the cabin is dimly lit by indoor lights.
For the counting-style question, vanilla misreads the fastest runner's bib number as \textcolor{red}{1019}, but our method outputs the correct number \textcolor{blue}{1097}, aligned with the ground-truth.
These examples support our quantitative findings that ACG improves informativeness while reducing hallucination on MMHal-Bench.

\subsection{Qualitative CHAIR, VQA Examples.}

\begin{figure*}[t]
    \centering
    \includegraphics[width=0.9\linewidth]{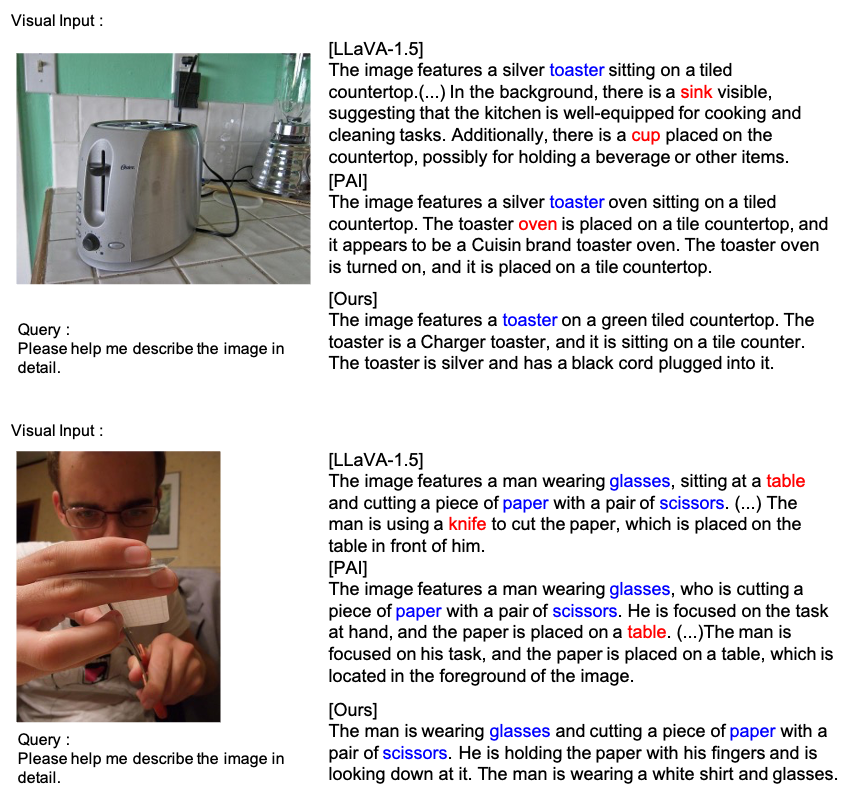}
    \caption{
    \textbf{Qualitative CHAIR examples comparing vanilla LLaVA-1.5, PAI, and our ACG method.}
    For each caption, we highlight object tokens: \textcolor{blue}{blue} tokens denote objects that are grounded in the image, while \textcolor{red}{red} tokens indicate hallucinated objects (e.g., \textcolor{red}{sink}, \textcolor{red}{cup}, \textcolor{red}{table}, \textcolor{red}{knife}).
    Compared to the baselines, ACG removes spurious background objects and focuses on the truly visible entities (e.g., the toaster, glasses, paper, and scissors).
    }
    \label{fig:chair_qual}
\end{figure*}

Figure~\ref{fig:chair_qual} shows CHAIR-style captioning examples.
In the toaster image (top), vanilla LLaVA-1.5 and the PAI baseline hallucinate background objects such as a \textcolor{red}{sink}, \textcolor{red}{cup}, or even misclassify the appliance as a \textcolor{red}{toaster oven}, while our method only mentions the grounded \textcolor{blue}{toaster} and its visible attributes.
In the second example (bottom), both baselines repeatedly refer to a \textcolor{red}{table} and a \textcolor{red}{knife} that are not clearly present in the image.
ACG instead concentrates on the truly visible entities---\textcolor{blue}{glasses}, \textcolor{blue}{paper}, and \textcolor{blue}{scissors}---showing that attention-space guidance effectively suppresses spurious background objects while preserving the core scene semantics. We also present qualitative results from MMMU and MathVista in Fig.~\ref{fig:mmmu_qual} and Fig.~\ref{fig:mathv_qual}.

%%%%%%%%%%Qualatative Examples-CHAIR

%%%%%%%%%%Qualatative Examples

\begin{figure}[t]
\centering
\includegraphics[width=0.7 \linewidth]{figure/mmmu_val_accounting_3.png}

\vspace{0.5em}

\textbf{Q:} Maxwell Software, Inc., has the following mutually exclusive projects.Suppose the company uses the NPV rule to rank these two projects. Which project should be chosen if the appropriate discount rate is 15 percent? \\

\vspace{0.3em}

\textbf{Choices :}  A : Project A , B : Project B \\ 
\textbf{GT:} B \\
\textbf{Baseline:} \textcolor{red}{A} \\
\textbf{Ours:} \textcolor{blue}{B}

\vspace{1.3em}
\centering
\includegraphics[width=1.0 \linewidth]{figure/mmmu_arch_eng_11.png}

\vspace{0.5em}

\textbf{Q:} The results of a compaction test on samples of soil that are to be used for an embankment on a highway project are listed below. Determine the optimum moisture content. \\

\vspace{0.3em}

\textbf{Choices:} A : 10\%. , B : 8\%. , C : 9\%. \\
\textbf{GT:} B \\
\textbf{Baseline:} \textcolor{red}{C} \\
\textbf{Ours:} \textcolor{blue}{B}

\caption{\textbf{Qualitative comparison on MMMU benchmark.}}
\label{fig:mmmu_qual}

\end{figure}

\begin{figure}[t]
\centering
\includegraphics[width=1.0 \linewidth]{figure/mathvista_25.jpg}

\vspace{0.5em}

\textbf{Q:} Is Medium Periwinkle the smoothest?
\vspace{0.3em}

\textbf{Choices :}  yes, no \\ 
\textbf{GT:} no \\
\textbf{Baseline:} \textcolor{red}{yes} \\
\textbf{Ours:} \textcolor{blue}{no} \\ 

\vspace{1.3em}
\centering
\includegraphics[width=0.7 \linewidth]{figure/mathv_109.jpg}

\vspace{0.5em}

\textbf{Q:} Subtract all tiny balls. Subtract all green metallic things. How many objects are left?\\

\vspace{0.3em}

\textbf{GT:} 5 \\
\textbf{Baseline:} \textcolor{red}{4} \\
\textbf{Ours:} \textcolor{blue}{5}

\caption{\textbf{Qualitative comparison on Mathvista benchmark.}}
\label{fig:mathv_qual}
\end{figure}

% {
%     \small
%     \bibliographystyle{ieeenat_fullname}
%     \bibliography{main}
% }
\end{document}